\documentclass[a4paper,fleqn]{cas-sc}

\usepackage[authoryear]{natbib}
\usepackage{subfigure}
\usepackage{algorithm}
\usepackage{algorithmicx}
\usepackage{algpseudocode}
\usepackage{caption}
\usepackage{xcolor}

\usepackage{amsthm}
\usepackage[section]{placeins} 
\usepackage{longtable}

\usepackage{enumitem}

\newlist{lightitemize}{itemize}{1}
\setlist[lightitemize]{
  label=\textbullet,
  leftmargin=\parindent,
  itemsep=0.15em,
  topsep=0.15em,
  parsep=0em,
  partopsep=0em
}

\newlist{lightdescription}{description}{1}
\setlist[lightdescription]{
  labelindent=\parindent,
  leftmargin=!,
  labelwidth=60 px,
  labelsep=0.5em,
  itemsep=0.5em,
  topsep=0.5em,
  parsep=0em,
  partopsep=0em,
  font=\normalfont\bfseries
}

\def\tsc#1{\csdef{#1}{\textsc{\lowercase{#1}}\xspace}}
\tsc{WGM}
\tsc{QE}
\tsc{EP}
\tsc{PMS}
\tsc{BEC}
\tsc{DE}

\hypersetup{
  colorlinks,
  citecolor=blue,
  linkcolor=blue,
  urlcolor=blue}
\begin{document}
\let\WriteBookmarks\relax
\def\floatpagepagefraction{1}
\def\textpagefraction{.001}

\shorttitle{V2I Work Zone Geometry Reconstruction with Pose-Conditioned UWB Range Denoising}
\shortauthors{Liu et al.}

\title[mode=title]{Vehicle-to-Infrastructure Work Zone Geometry Reconstruction with Pose-Conditioned UWB Range Denoising}

\author[1]{Jiaxi Liu}

\author[1]{Hangyu Li}[orcid=0000-0002-8667-9928]
\cormark[1]
\ead{hangyu.li@wisc.edu}

\author[1]{Yang Cheng}

\author[1]{Rui Gan}

\author[1]{Junwei You}

\author[1]{Weizhe Tang}

\author[1]{Peng Zhang}

\author[1]{Steven T. Parker}

\author[1]{Xiaopeng Li}[orcid=0000-0002-5264-3775]

\author[1]{Bin Ran}

\affiliation[1]{organization={Department of Civil \& Environmental Engineering, University of Wisconsin-Madison},
    city={Madison},
    postcode={53706},
    state={Wisconsin},
    country={United States}}

\cortext[cor1]{Corresponding author}

\begin{abstract}
Reliable work zone mapping is important for connected and autonomous vehicles (CAVs) to navigate safely and smoothly through work zone areas. Cone-mounted ultra-wideband (UWB) roadside units (RSU) offer a cost-effective way for work zone layout inference, as roadside anchors and vehicle tags provide direct vehicle-to-infrastructure (V2I) range constraints for work zone geometry reconstruction. However, UWB range estimation is degraded by bursty outliers, non-line-of-sight (NLOS) errors, arbitrary anchor-ordering issues, and vehicle pose uncertainties in practical field deployments. 
To address these challenges, this study proposes a pose-conditioned, permutation-equivariant predictive denoiser for multi-anchor UWB ranging. The model employs shared anchor-wise temporal prediction to capture range dynamics, symmetric set aggregation to handle unordered and missing anchors, and pose-conditioned residual decoding to incorporate vehicle motion as a geometric prior. A two-stage training strategy first learns prediction from observed ranges, and then fine-tunes the denoiser with NLOS-weighted supervision.  The method is evaluated on rare real-world V2I UWB field data collected with a CAV, as well as on controlled large-scale simulation benchmarks for ablative insights. Results show that the proposed method substantially improves range accuracy, cone localization, and work zone geometry reconstruction in challenging NLOS-dominated regimes, remains robust to anchor re-indexing and moderate anchor dropout, and reduces measurement-weighted field MSE by 66.9\% relative to the raw input.
\end{abstract}

\begin{highlights}
\item \textbf{V2I UWB work zone mapping}: Cone-mounted UWB devices are formulated as roadside anchors that cooperate with vehicle tags to provide direct range constraints for cone localization and work zone geometry reconstruction.

\item \textbf{Pose-conditioned equivariant range denoising}: A denoiser model combines anchor-wise temporal prediction, set-level aggregation, pose-conditioned residual decoding, and a predictive pretraining plus fine-tuning strategy.

\item \textbf{Field and simulation validation}: Real-world V2I experiments and controlled simulations show improved range accuracy, NLOS robustness, anchor re-indexing and dropout stability, and downstream work zone reconstruction.
\end{highlights}


\maketitle

\section{Introduction}
\label{sec:intro}

Reliable work zone mapping is important for connected and autonomous vehicles (CAVs) and advanced traffic management systems because work zone geometry and traffic-control devices can change rapidly. Lane shifts, tapers, narrowed shoulders, cone lines, and temporary barriers may be installed, moved, or removed within hours or days, posing challenges for CAVs with uncertainties \citep{li2025robotic}. As a result, CAVs require an up-to-date representation of the active work area and the remaining drivable region to navigate safely and smoothly through work zones \citep{dehman2021workzone_cav_review}. When visibility is limited or when the layout changes faster than onboard perception can update, vehicle-only sensing may provide an incomplete picture. Vehicle-to-infrastructure (V2I) sensing can help address this limitation by allowing roadside units to provide direct geometric information to passing vehicles \citep{liu2024adaptive}. Recent work zone data initiatives, such as the Work Zone Data Exchange (WZDx), further highlight the need for harmonized and machine-readable work zone information that can be consumed by navigation systems, original equipment manufacturers, and automated driving systems \citep{usdot2024wzdx}. However, static work zone data feeds do not provide dynamic map descriptions observed by a passing vehicle, which motivates real-time cone localization and precise work zone geometry reconstruction.

Existing work zone mapping pipelines mainly rely on three types of information. First, survey-grade or infrastructure-based mobile mapping systems can recover detailed roadway geometry, such as lane-width and cross-section changes, from calibrated sensors (e.g., LiDAR) \citep{habib2018lidar_lane_width_workzones}. Second, vision-based approaches detect temporary traffic control devices (TTCDs) in work zones, such as cones, barrels, signs, workers, and equipment, and then infer work zone extent through scene understanding and topology-based reasoning \citep{seo2022ttcd_crc,zuo2023urban_workzone_detection_sizing,shi2021work}. Third, trajectory-based approaches use crowd-sourced vehicle trajectories to infer temporarily changed drivable areas and support downstream navigation \citep{chen2023crowdsourcing_workzone_mapping}. These approaches are valuable but have complementary limitations. Mapping-grade systems provide accurate geometry but require dedicated equipment and calibration. Vision-based methods are sensitive to occlusion, lighting, and object appearance variation, as well as limited by annotation effort, remaining a long-tailed problem \citep{ghosh2025roadwork}. Trajectory-based methods may respond slowly to abrupt layout changes or become unreliable when vehicle observations are sparse. A temporary work zone sensing system should therefore be low-cost, rapidly deployable, less dependent on visual scene understanding alone, and able to provide direct metric constraints as the layout changes.

Ultra-wideband (UWB) ranging is promising for this role because it provides direct distance measurements with fine time resolution and can be deployed with low-cost roadside hardware \citep{gezici2005uwb_localization_spm,alarifi2016uwb_indoor_review,volpi2023lowcost_uwb_rtls}. UWB has also been used for tracking and hazard zone monitoring in construction and roadwork safety applications \citep{maalek2016uwb_construction_accuracy,ochoa2024uwb_roadworker_safety}. In a V2I work zone mapping system, UWB-enabled roadside units (UWB-RSUs) can be mounted on traffic cones and treated as temporary roadside anchors. A vehicle-mounted UWB tag then measures ranges to these anchors while the vehicle passes through or near the work zone, turning the cone layout into a set of constraints for cone localization and work zone geometry reconstruction. In this way, UWB does not replace survey-grade, vision-based, or trajectory-based mapping, but provides a complementary low-cost ranging layer for temporary work zone deployments.

The remaining challenge is that field UWB ranges must be reliable enough to support this downstream mapping task. Non-line-of-sight (NLOS) propagation and multipath can introduce positive range bias, while transient interference can create burst outliers and heavy-tailed residuals \citep{wang2023uwb_nlos_survey,angarano2021robust_uwb_deep_edge}. Classical filtering and robust state-space methods, including adaptive Kalman filtering for time-of-flight systems, provide interpretable and low-latency baselines \citep{qiyue2015adaptive_kf_nlos}. However, simple per-step filters may be less reliable when NLOS errors persist over multiple time steps or when large impulses appear in short bursts. Learning-based UWB error mitigation can capture richer nonlinear corrections, especially when physical-layer signals or temporal context are available \citep{angarano2021robust_uwb_deep_edge,niu2023deep,yang2024ultra}. At the same time, purely supervised learning remains challenging because accurate ground-truth distances require survey-grade references that are expensive to collect at scale. Semi-supervised and predictive learning strategies therefore offer a useful direction by leveraging abundant unlabeled range streams before supervised calibration \citep{wang2021semi_supervised_uwb_wcl,li2023semi_supervised_uwb_waveform,assran2023ijepa}. These limitations motivate a denoising method that first learns temporal regularity from observed range sequences and then uses available ground-truth distances for supervised refinement.

In addition to range noise, a V2I work zone mapping system must handle changes in the anchor set. The number of visible UWB-RSUs can vary across deployments and across time because of occlusion, equipment movement, temporary blockage, or operational changes. Moreover, raw ranging interfaces may return visible anchors as unordered anchor-range pairs, so the same physical anchors can appear under different input orders after software-level re-indexing. A fixed-order vector representation can therefore become fragile when the input order changes or when some anchors are missing. Set-structured learning provides a natural way to address this issue through shared element-wise transformations and symmetric aggregation, as formalized in Deep Sets and extended in attention-based set models \citep{zaheer2017deepsets,lee2019set_transformer}. For work zone UWB ranging, this means that the denoiser should treat the anchor dimension as a masked set rather than as a fixed-order vector. Robustness to anchor re-indexing and anchor dropout is therefore a practical deployment requirement rather than only a modeling preference.

To address these challenges, this paper studies UWB-assisted work zone mapping as a V2I geometry-inference problem. The central idea is to improve the reliability of multi-anchor UWB ranges before they are used for cone localization and work zone geometry reconstruction. We propose a pose-conditioned, permutation-equivariant predictive denoiser for multi-anchor UWB ranging. The model combines shared anchor-wise temporal prediction to capture range dynamics, symmetric set aggregation to handle unordered and missing anchors, and pose-conditioned residual decoding to use the vehicle pose stream as a geometric prior. The model is trained in two stages: predictive pretraining on observed ranges followed by NLOS-weighted supervised fine-tuning. This design aims to improve range accuracy under bursty NLOS errors while maintaining consistency under anchor re-indexing and dropout.

Our contributions are threefold:
\begin{lightitemize}
  \item We formulate UWB-assisted work zone mapping as a V2I geometry-inference problem, in which cone-mounted UWB-RSUs serve as anchors and multi-anchor range measurements provide direct constraints for cone localization and work zone geometry reconstruction.
  \item We develop a pose-conditioned, permutation-equivariant predictive denoiser for multi-anchor UWB ranging. The model combines anchor-wise temporal prediction, symmetric set aggregation, and pose-conditioned residual decoding to cope with burst outliers and NLOS errors while handling unordered anchor inputs and missing anchors.
  \item We validate the proposed method on both measured V2I field data and large-scale controlled simulations, evaluating range denoising, cone localization, work zone geometry reconstruction, anchor re-indexing and dropout robustness, component sensitivity, and pose-noise sensitivity.
\end{lightitemize}

\begin{figure}[]
    \centering
    \includegraphics[width=0.6\linewidth]{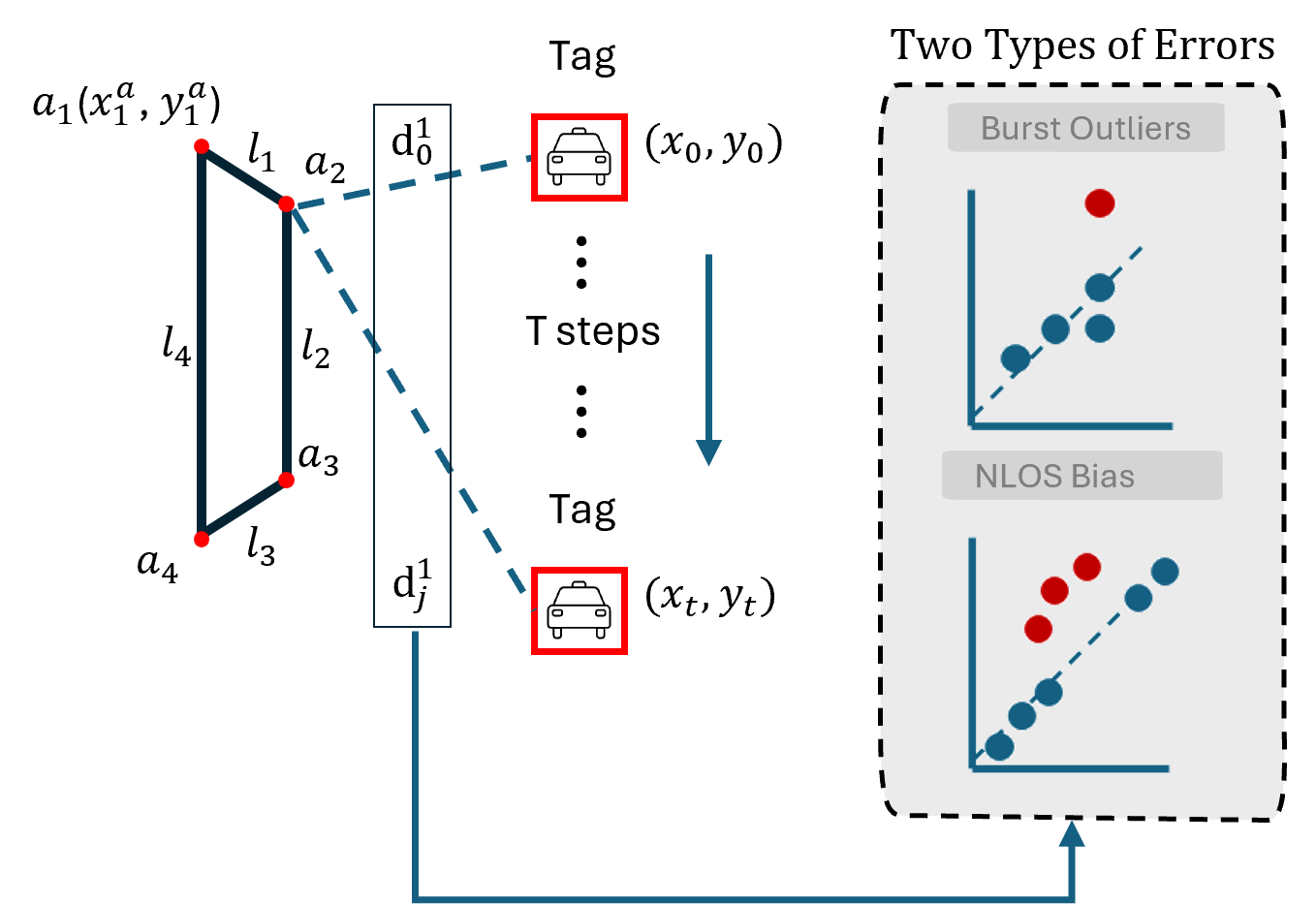}
    \caption{Work zone UWB ranging scenario and the two dominant ranging error modes: burst outliers (isolated high-amplitude deviations) and NLOS bias (sustained positive or negative shift).}
  \label{fig:scenario_error_modes}
\end{figure}

\section{Problem Setup}
\label{sec:meas_sym_downstream}

\subsection{V2I Sensing Scenario, Data Representation, and Task Definition}
We consider a V2I-enabled sensing system in which UWB-RSUs provide infrastructure-side ranges from work zone cones to a moving vehicle-mounted tag. Each UWB-RSU is a UWB anchor mounted on a traffic cone and treated as a single fixed infrastructure-side ranging anchor. Figure~\ref{fig:scenario_error_modes} shows the sensing geometry together with the two error patterns that motivate the formulation below. Burst outliers are isolated deviations caused by transient interference or abrupt multipath events. NLOS bias is a sustained positive shift that appears when the direct ranging path is blocked and reflections dominate. These two effects have distinct temporal signatures, so the method later combines temporal prediction with bias-aware method rather than treating all errors as independent, identically distributed noise.

A work zone episode is a trajectory segment of length $T$ in which the vehicle-mounted tag measures ranges to a set of UWB-RSUs. We index episodes by $e\in\{1,\dots,E\}$, time steps by $t\in\{1,\dots,T\}$, and anchor slots by $i\in\{1,\dots,N_{\max}\}$; each episode is right-padded to $N_{\max}$ for batch processing, and the structural mask separates real anchor slots from padded ones. We write $\mathcal{T}=\{1,\dots,T\}$ and $\mathcal{N}=\{1,\dots,N_{\max}\}$ for the time and padded anchor-slot index sets. Anchor identity is fixed within an episode in our benchmarks, and sustained anchor unavailability is evaluated separately through anchor-dropout stress tests.

For a fixed episode and indices $t,i$, the raw inputs are the vehicle-mounted tag position $p_t \in \mathbb{R}^2$, the observed raw range $d^{obs}_{t,i}$, and the structural validity mask $m_{t,i}\in\{0,1\}$.

We additionally have ground-truth ranges $d^{gt}_{t,i}$ and an NLOS indicator $n_{t,i}\in\{0,1\}$ for supervised training and diagnostics during training phase. Collecting these objects at the episode level gives:
\begin{equation}
\mathbf{P}=\{p_t\}_{t\in\mathcal{T}},\quad
\mathbf{D}^{obs}=\{d_{t,i}^{obs}\}_{t\in\mathcal{T},\, i\in\mathcal{N}},\quad
\mathbf{M}=\{m_{t,i}\}_{t\in\mathcal{T},\, i\in\mathcal{N}}.
\end{equation}

The denoiser $f_\theta$ maps these sequence inputs to corrected ranges:
\begin{equation}
\hat d_{t,i} = f_\theta\!\left(\mathbf{P}, \mathbf{D}^{obs}, \mathbf{M}\right)_{t,i}.
\end{equation}

The goal in this range phase is to minimize overall, line-of-sight (LOS), and NLOS range error for UWB-enabled distance estimation and also preserve prediction consistency under episode-consistent anchor permutation and anchor dropout situation.


\subsection{UWB Observation Model and Error Structure}
\label{sec:meas_model}

Let $a_i=(x_i^a,y_i^a)\in\mathbb{R}^2$ denote the 2D coordinate of the infrastructure anchor associated with slot $i$, and write the reference tag position as $p_t=(x_t,y_t)\in\mathbb{R}^2$. The ground-truth distance is:
\begin{equation}
d^{gt}_{t,i} = \|p_t - a_i\|_2 .
\end{equation}

\paragraph{Anchor identity and episode alignment.}
Throughout, the anchor index $i$ denotes an anchor identity that is consistent across time within an episode (e.g., a stationary UWB-RSU broadcasting a unique ID, or an anchor with a known fixed coordinate that can be matched by proximity). In practice, raw ranging APIs often return an unordered list of anchor id and raw measured distance pairs whose list order may vary across calls. a lightweight normalized step aligns these measurements into a fixed index order. Under this alignment, software-induced order changes are well modeled as an episode-consistent permutation, which Section~\ref{sec:peq_equivariance} formalizes as the structural property the denoiser must satisfy. 

The observed range is modeled as:
\begin{equation}
d^{obs}_{t,i} =
m_{t,i}\Big(d^{gt}_{t,i} + b_{t,i} + \beta^+_{t,i} + \varepsilon_{t,i}\Big),
\end{equation}
where $b_{t,i}$ denotes signed per-anchor offset and drift, $\beta^+_{t,i}\ge 0$ denotes NLOS inflation, and $\varepsilon_{t,i}$ collects zero-mean transient perturbations, including temporally correlated noise and occasional signed heavy-tailed impulses. This decomposition separates signed anchor-specific calibration error, sustained positive propagation inflation, and transient perturbations. In Figure~\ref{fig:scenario_error_modes}, burst outliers mainly appear as large transient realizations of $\varepsilon_{t,i}$, whereas NLOS segments are characterized by sustained positive $\beta^+_{t,i}$ together with heavier-tailed residual variation. For simulation analysis, an indicator $n_{t,i}$ marks whether a sample is NLOS-affected. It is used for diagnostics and loss weighting.

The base LOS measurement component is modeled as zero-mean Gaussian noise, but the benchmark also adds per-anchor bias and drift and occasional heavy-tailed impulses, so LOS-labeled samples can still show non-Gaussian residual tails. In the NLOS state ($n_{t,i}=1$), additional positive, time-varying bias terms become active. Section~\ref{sec:data_gen} instantiates these states with concrete parameters for different simulation regimes.

The denoiser takes $p_t$ as an auxiliary input to leverage motion regularity and geometry context. In deployment, $p_t$ will come from RTK/INS. To avoid unrealistic information leakage, we later evaluate injected pose perturbations at test time rather than assuming perfect pose.

\section{Proposed Pose-Conditioned Permutation-Equivariant Predictive Denoiser}

\begin{figure}
  \centering
  \includegraphics[width=0.98\linewidth]{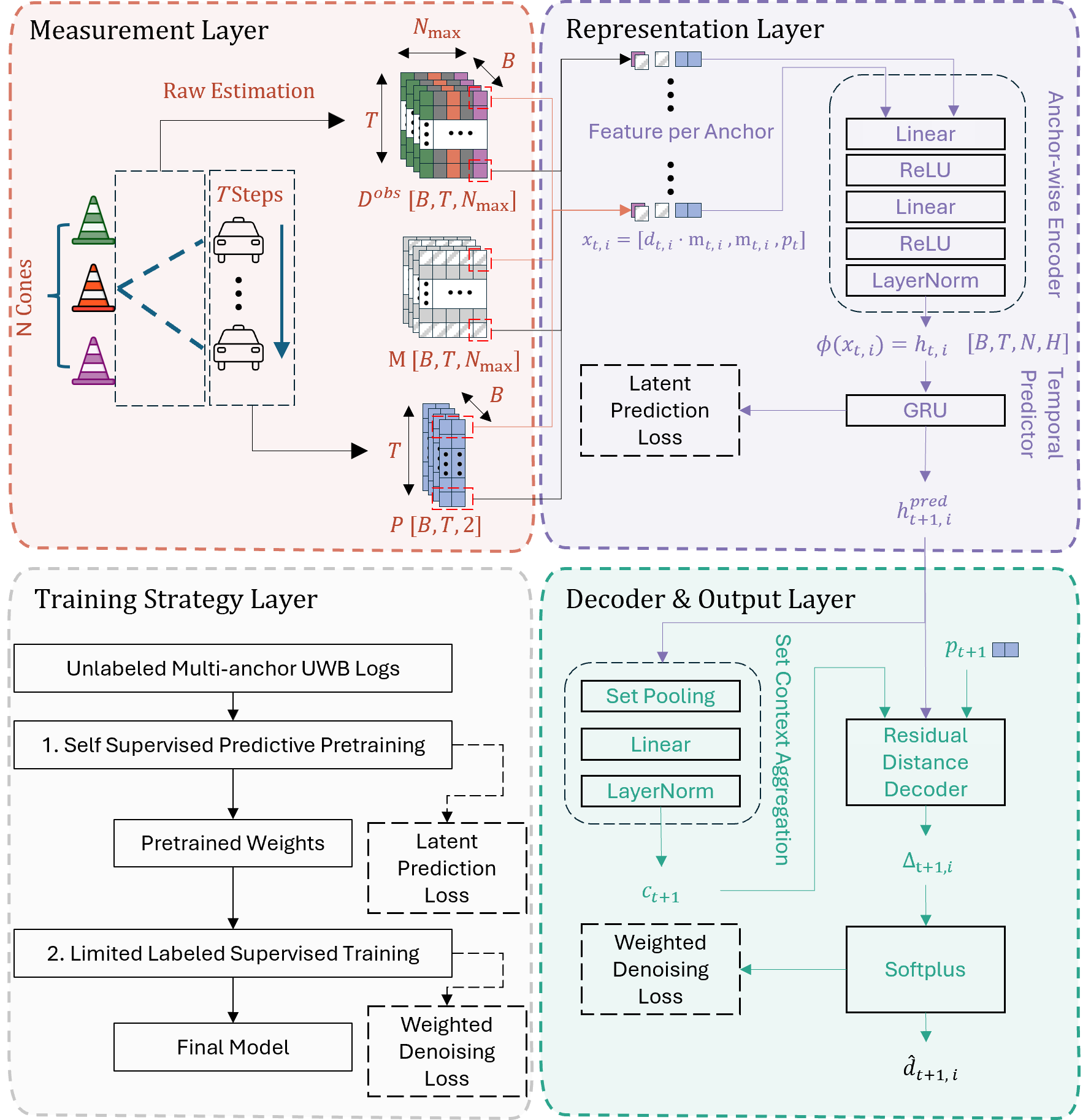}
  \caption{Overall pipeline of the V2I-enabled inference backbone: the measurement layer aligns raw UWB logs into pose--range--mask sequences. The representation layer applies shared anchor-wise encoding and temporal prediction. The decoder aggregates set-level context and outputs non-negative denoised distances through residual correction. Training proceeds in two stages (predictive pretraining followed by NLOS-aware supervised fine-tuning). Here $B$ denotes mini-batch size, $T$ sequence length, $N_{\max}$ padded anchor slots, and $H$ latent width, $i$ indexes anchor slots, while $n_{t,i}$ is reserved for the simulation-only NLOS indicator.}
  \label{fig:overall_framework}
\end{figure}

Figure~\ref{fig:overall_framework} shows the proposed pipeline. The architecture combines four components: \emph{shared anchor-wise temporal modeling} for bursty range errors as shown in the \textbf{Representation Layer}, \emph{symmetric set-level aggregation} for unordered and variable-cardinality anchor observations as shown in the \textbf{Decoder \& Output Layer}, pose-conditioned residual decoding for geometry-aware correction as shown in the \textbf{Decoder \& Output Layer}, and a \emph{two-stage training protocol} with predictive pretraining followed by NLOS-aware supervised fine-tuning as shown in the \textbf{Training strategy layer}. The remainder of this section specifies architecture of the backbone, training objectives, equivariance property, and inference behavior.

\subsection{Architecture of the Predictive Denoiser}
\label{sec:peq_denoiser}
The model takes $\mathbf{P}$, $\mathbf{D}^{obs}$, and $\mathbf{M}$ as inputs. At each step it first forms a predictive latent prior for time $t{+}1$ from the history up to $t$, then combines this prior with the same-step raw measurement through residual decoding. The backbone outputs $\hat d_{t+1,i}$ together with predictive latents $\hat h_{t+1,i}$ used during training. When referring to the aligned denoised sequence generically, we write $\hat d_{t,i}$. The validity mask gates padded anchor slots in the forward pass, while the NLOS indicator $n_{t,i}$ is used only in supervised loss weighting during training and never as a denoiser input at inference time.

\textbf{Step 1: Local anchor-state encoding.}
Let $\phi(\cdot)$ denote a shared multilayer perceptron (MLP) applied independently to each anchor stream. We build local features as:
\begin{equation}
h_{t,i} = \phi\left([d^{obs}_{t,i}m_{t,i},\; m_{t,i},\; p_t]\right).
\end{equation}

Including $p_t$ in the per-anchor input allows the encoder to condition each range stream on the local vehicle state from the first layer.

\textbf{Step 2: Shared temporal prediction.}
Let $\mathrm{GRU}(\cdot)$ denote the shared gated recurrent unit (GRU) used as the temporal predictor. Anchor-wise temporal prediction is performed by:
\begin{equation}
\hat h_{t+1,i} = \mathrm{GRU}(h_{1:t,i}).
\end{equation}

Using one shared predictor across anchors encourages consistent dynamics modeling and reduces anchor-ID-specific overfitting. By accumulating recurrent state along each anchor's history, this step captures temporal structure that is difficult for per-step filters or memoryless feed-forward learners to exploit under bursty, non-Gaussian range errors.

\textbf{Step 3: Set-level context aggregation.}
Let $\rho(\cdot)$ denote a shared context MLP, and let $\varepsilon_0>0$ denote a small numerical-stability constant. Set-level context at each step is pooled from predicted latents:
\begin{equation}
c_{t+1} = \rho\left(\frac{\sum_i m_{t+1,i}\hat h_{t+1,i}}{\sum_i m_{t+1,i}+\varepsilon_0}\right).
\end{equation}

This context provides cross-anchor coupling while remaining permutation invariant through symmetric averaging. Combined with the shared anchor-wise operators, this design yields equivariance to anchor re-indexing and provides a mask-aware interface for the dropout stress tests.

\textbf{Step 4: Residual distance decoding.}
Let $g(\cdot)$ denote the shared residual decoder. Residual decoding predicts a correction to the same-step raw range:
\begin{subequations}
\begin{flalign}
\Delta_{t+1,i} &= g([\hat h_{t+1,i}, c_{t+1}, p_{t+1}]),\\
\hat d_{t+1,i} &= \text{softplus}(d^{obs}_{t+1,i} + \Delta_{t+1,i})m_{t+1,i}.
\end{flalign}
\end{subequations}

The decoder uses the predictive latent state and the pose-aware set context as priors, correcting the same-step observation rather than regressing an absolute distance from scratch. The \texttt{softplus} nonlinearity enforces non-negative range outputs. Pose is therefore used both in anchor-wise encoding and residual decoding, allowing the network to condition corrections on local motion and global geometry.

\subsection{Training Objectives and Two-Stage Learning}

\textbf{Stage 1 (unlabeled predictive pretraining).}
Let $\text{sg}(\cdot)$ denote the stop-gradient operator, which blocks gradient flow through its argument so that the predictor is trained toward a fixed latent target. The first phase uses unlabeled range logs to learn temporal regularity and cross-anchor latent structure:
\begin{subequations}
\begin{flalign}
&\mathcal{L}_{pred} = \text{MaskedMSE}(\hat h_{t+1,i}, \text{sg}(h_{t+1,i})), \\
&\mathcal{L}_{dec}^{obs} = \text{MaskedHuber}(\hat d_{t+1,i}, d^{obs}_{t+1,i}).
\end{flalign}

The Stage 1 objective is:
\begin{equation}
\mathcal{L}_{stage1} = \mathcal{L}_{pred} + \lambda_{obs}\mathcal{L}_{dec}^{obs},
\end{equation}
\end{subequations}
where $\lambda_{obs}\ge 0$ controls the strength of the observation-space auxiliary term. The masked losses are evaluated only over valid anchor slots, and the Huber penalty keeps fitting robust to heavy-tailed outliers, which is quadratic for small residuals, linear for large ones. This stage teaches the model to forecast temporally coherent latent evolution and to decode measurements without requiring ground-truth distance labels. The anchor-wise encoder, shared GRU predictor, set-level context module, and residual decoder are optimized jointly.

\textbf{Stage 2 (supervised denoising fine-tuning).}
The second phase switches targets from observed range to ground truth:
\begin{subequations}  
\begin{flalign}
&w^{sup}_{t+1,i}=m_{t+1,i}\left[(1-n_{t+1,i})+\alpha n_{t+1,i}\right],\quad 0<\alpha<1, \\
&\mathcal{L}_{sup}=\text{WeightedHuber}(\hat d_{t+1,i}, d^{gt}_{t+1,i}; w^{sup}_{t+1,i}).
\end{flalign}

If predictive regularization is retained during fine-tuning, the Stage 2 objective becomes:
\begin{equation}
\mathcal{L}_{stage2} = \mathcal{L}_{sup} + \lambda_{pred}\mathcal{L}_{pred},
\end{equation}
\end{subequations}
where $\lambda_{pred}\ge 0$ controls the extent to which the predictive prior is preserved after supervised calibration. Weighted supervision avoids over-penalizing difficult NLOS samples while still learning from them.

\paragraph{Design role of the two stages.}
The two stages play complementary roles in the architecture: Stage 1 supplies an initialization that already captures cross-anchor temporal regularity from raw observations, and Stage 2 then refines the same backbone toward the ground-truth distance target under NLOS-aware weighting. The empirical contribution of this Stage 1 initialization is quantified by the ``w/o pretraining'' row of the component ablation in Section~\ref{sec:robust}.

\subsection{Equivariance Property of the Proposed Architecture}
\label{sec:peq_equivariance}

We first state the structural requirement and then show the architecture satisfies it by construction. Let $S_{N_{\max}}$ denote the symmetric group on anchor slots, and let $\pi\in S_{N_{\max}}$ act on the padded range and mask tensors by $(\pi\mathbf{D})_{t,i}=\mathbf{D}_{t,\pi^{-1}(i)}$ and $(\pi\mathbf{M})_{t,i}=\mathbf{M}_{t,\pi^{-1}(i)}$. A denoiser $f_\theta$ that outputs anchor-wise corrected ranges is \emph{permutation equivariant} if:
\begin{equation}
f_\theta(\mathbf{P}, \pi\mathbf{D}^{obs}, \pi\mathbf{M}) = \pi f_\theta(\mathbf{P}, \mathbf{D}^{obs}, \mathbf{M}), \quad \forall \pi \in S_{N_{\max}}.
\label{eq:equiv_def}
\end{equation}

This covers episode-consistent re-indexing, such as sorting by ID, by coordinate, or by an arbitrary API list order. The same masking interface lets anchor dropout be evaluated by zeroing selected anchor masks, but dropout robustness is assessed empirically rather than implied by equivariance alone. Time-varying permutations $\{\pi_t\}_{t=1}^T$, which represent data-association failures, are out of scope.

The proposed architecture satisfies Eq.~\eqref{eq:equiv_def} by construction. The anchor-wise encoder $\phi(\cdot)$, temporal predictor $\mathrm{GRU}(\cdot)$, and residual decoder $g(\cdot)$ share parameters across anchors and are applied independently to each anchor stream, so re-indexing the inputs by $\pi$ re-indexes their latent and output sequences in exactly the same way. The set-level context is a symmetric (masked-mean) aggregation and is therefore invariant to anchor order. Composing these steps gives the equivariance of Eq.~\eqref{eq:equiv_def} for any $\pi\in S_{N_{\max}}$.

\subsection{Inference Procedure and Computational Characteristics}

At inference time only $(p_t, d^{obs}_{t,i}, m_{t,i})$ are needed. The model runs a single forward pass over the context window and feeds directly into the downstream solvers of Section~\ref{sec:downstream_eval}. The dominant complexity is $\mathcal{O}(BTN_{\max} H^2)$ for the shared GRU plus $\mathcal{O}(BTN_{\max} H)$ for the shared MLP encode/decode, where $B$ is batch size and $H$ hidden width. Because the anchor operators are shared and independent before pooling, the architecture scales linearly in anchor count for fixed hidden width. These design choices make the model lightweight at inference time.

\section{Experimental Setup}
\label{sec:exp_setup}

This section specifies how the formulation in Section~\ref{sec:meas_sym_downstream} and the method in Section~\ref{sec:peq_denoiser} are deployed in data, baselines, downstream solvers, and model-selection rules. Evaluation follows two complementary tracks: deployment-facing real-world V2I experiments and controlled large-scale simulation benchmarks.

\subsection{Real-World V2I-Enabled Experiments}

\subsubsection{System setup and geometry instantiation.}
\begin{figure}[]
  \centering
  \includegraphics[width=0.70\linewidth]{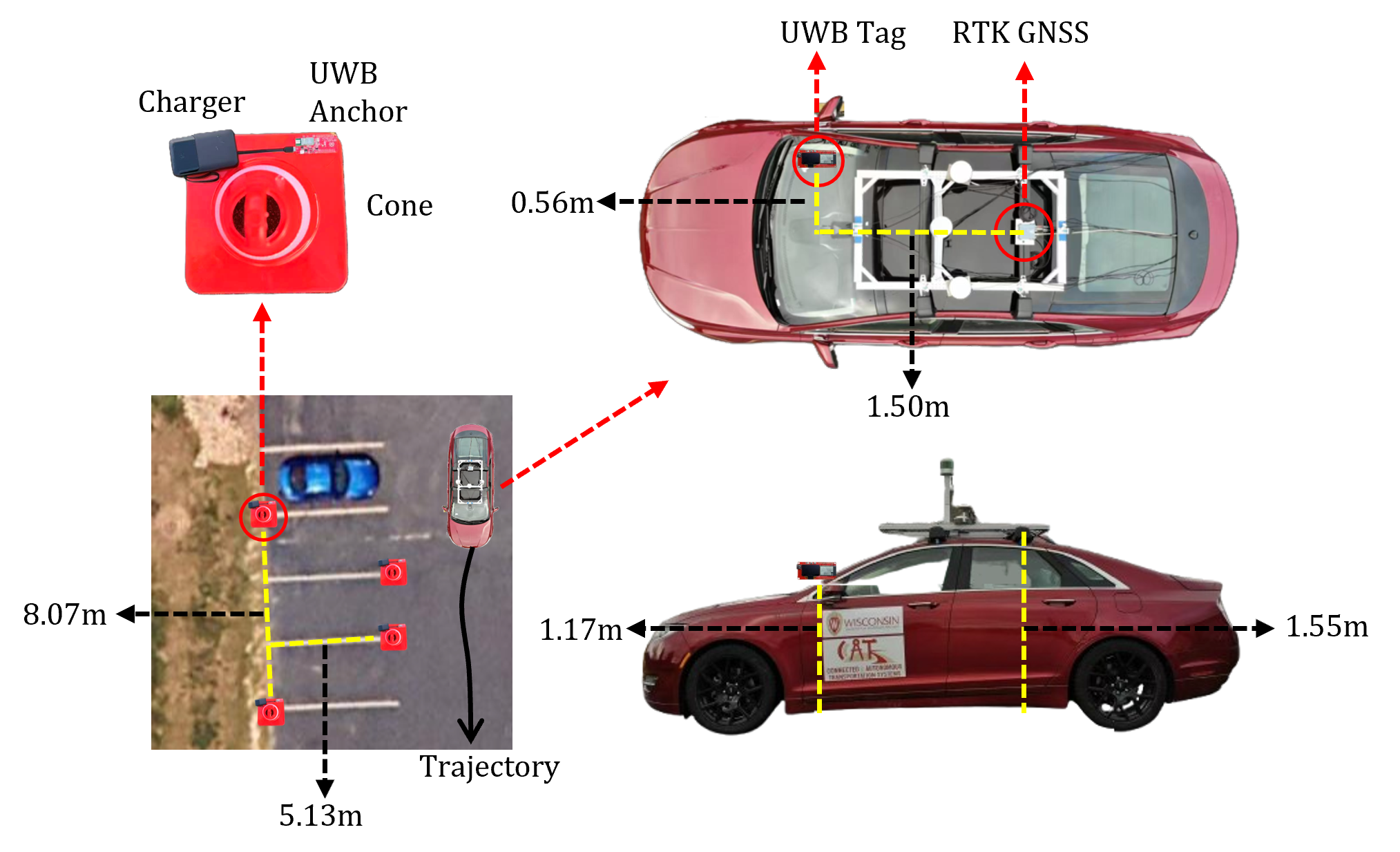}
  \caption{Real-world V2I experiment setup: cone-mounted UWB-RSU (top-left), in-vehicle UWB-tag/RTK-GNSS mounting and horizontal lever arm (top-right), tag and GNSS heights above ground (bottom-right), and overhead site photo with the trapezoidal anchor layout and driven trajectory (bottom-left).}
  \label{fig:real_setup}
\end{figure}

The real dataset contains four infrastructure-side UWB-RSUs. Each UWB-RSU is mounted on a traffic cone and treated as one fixed anchor. In the current field dataset, the anchor layout is instantiated from the measured site dimensions as a trapezoidal work zone proxy with short base 2.69 m, long base 8.07 m, and height 5.13m, together with an explicit anchor-ID-to-corner assignment. The vehicle carries a UWB tag and an RTK/INS unit, with the global navigation satellite system (GNSS) antenna $1.55$\,m above ground. The UWB tag is mounted at $1.17$\,m above ground and offset from the GNSS antenna by $1.50$\,m forward and $0.56$\,m to the right. The reference tag trajectory is obtained by converting the RTK/INS latitude--longitude stream to a local planar frame and applying the measured rigid GPS-to-tag offset using GPS azimuth, so that the tag position used by the denoiser and the geometry-based range reference are spatially aligned. Because the UWB tag and the cone-mounted UWB-RSUs sit at different heights, the geometry-derived field reference distance is computed as the 3D slant range $d^{ref}_{t,i}=\sqrt{d_{xy,t,i}^2+(h_{tag}-h_{anchor})^2}$, with $h_{tag}=1.17$\,m. The current field reference uses a ground-anchor approximation $h_{anchor}=0$, reported explicitly as part of the calibrated field diagnostic. This slant reference matches the physical UWB range, whereas the downstream solvers of Section~\ref{sec:downstream_eval} use a 2D planar range model. The resulting tag-height frame difference applies identically to all methods and therefore does not affect the relative field comparison. Figure~\ref{fig:real_setup} summarizes the field setup, including the cone-mounted UWB-RSU hardware, the in-vehicle UWB-tag/RTK-GNSS mounting and lever arm, and the trapezoidal anchor layout with the driven trajectory overlaid on the site plan.

\subsubsection{Data collection and preprocessing.}
Raw UWB events are aggregated into $0.10$\,s bins. Within each bin, the most recent measurement from each visible anchor is retained and missing anchors are left masked out. The dynamic reference trajectory is obtained from the RTK/INS stream after timestamp alignment and interpolation to the UWB event times. For each dynamic episode, a geometry-derived reference distance is then computed from the tag trajectory and the instantiated anchor layout. This yields a real-data sequence representation that matches the simulation-side interface: pose sequence, observed-range sequence, and validity mask sequence.

\subsubsection{Evaluation protocol on real data.}
The field track keeps the same family of denoisers as in simulation: raw input, Kalman (1D), MLP, PoseMLP, PoseKalman, and the proposed method as introduced in Section~\ref{sub:baselines}. The real-data analysis answers two questions. First, does the denoiser reduce signal-level ranging error relative to the geometry-derived reference distance? Second, do the denoised ranges remain useful for downstream localization and work zone geometry reconstruction under the same shared solvers used in simulation? The field reference distances come from measured site geometry plus RTK/INS-based tag pose, so the absolute downstream scale is bounded by the small site footprint. The relative ordering among the shared-input-calibrated pipelines is directly comparable. The proposed field-calibrated row is additionally reported as a calibrated diagnostic because it uses a degree-2 output calibration fitted to field reference distances.

\subsection{Simulation Benchmarks for Controlled Large-Scale Evaluation}
\label{sec:data_gen}

The simulation track instantiates the generic observation model of Section~\ref{sec:meas_model} in two regimes: the challenging scenario (burst NLOS, heavy-tailed outliers) with $1440$ episodes and $T{=}250$ steps, and the clean scenario (lower-noise, no burst corruptions) with $800$ episodes and $T{=}220$ steps. Both run at $50$\,Hz. Each episode samples an anchor count $N\sim\mathrm{UnifInt}[4,12]$, anchor positions uniformly over a $60\,\mathrm{m}\times 60\,\mathrm{m}$ area, and a tag trajectory generated by a bounded random walk in speed and heading starting at $p_0=(0,0)$. One anchor layout is reused across multiple trajectories. The observed range $d^{obs}_{t,i}$ adds six terms on top of the ground-truth distance: heteroscedastic (distance-dependent), first-order autoregressive (AR(1)) LOS noise, a per-anchor bias with drift, signed Laplace outliers, a positive NLOS-burst bias with geometric dwell time, a separate outage-burst corruption, and an offset for anchors outside the front-facing field of view (FOV). The challenging scenario uses all six terms and the clean scenario zeros out the two burst terms and uses a wider FOV, so the clean regime is a lower-noise negative control rather than an NLOS-free or purely Gaussian setting: per-anchor bias and drift, occasional outliers, and mild FOV-driven NLOS labels remain.

\begin{figure}[]
  \centering
  \includegraphics[width=0.98\linewidth]{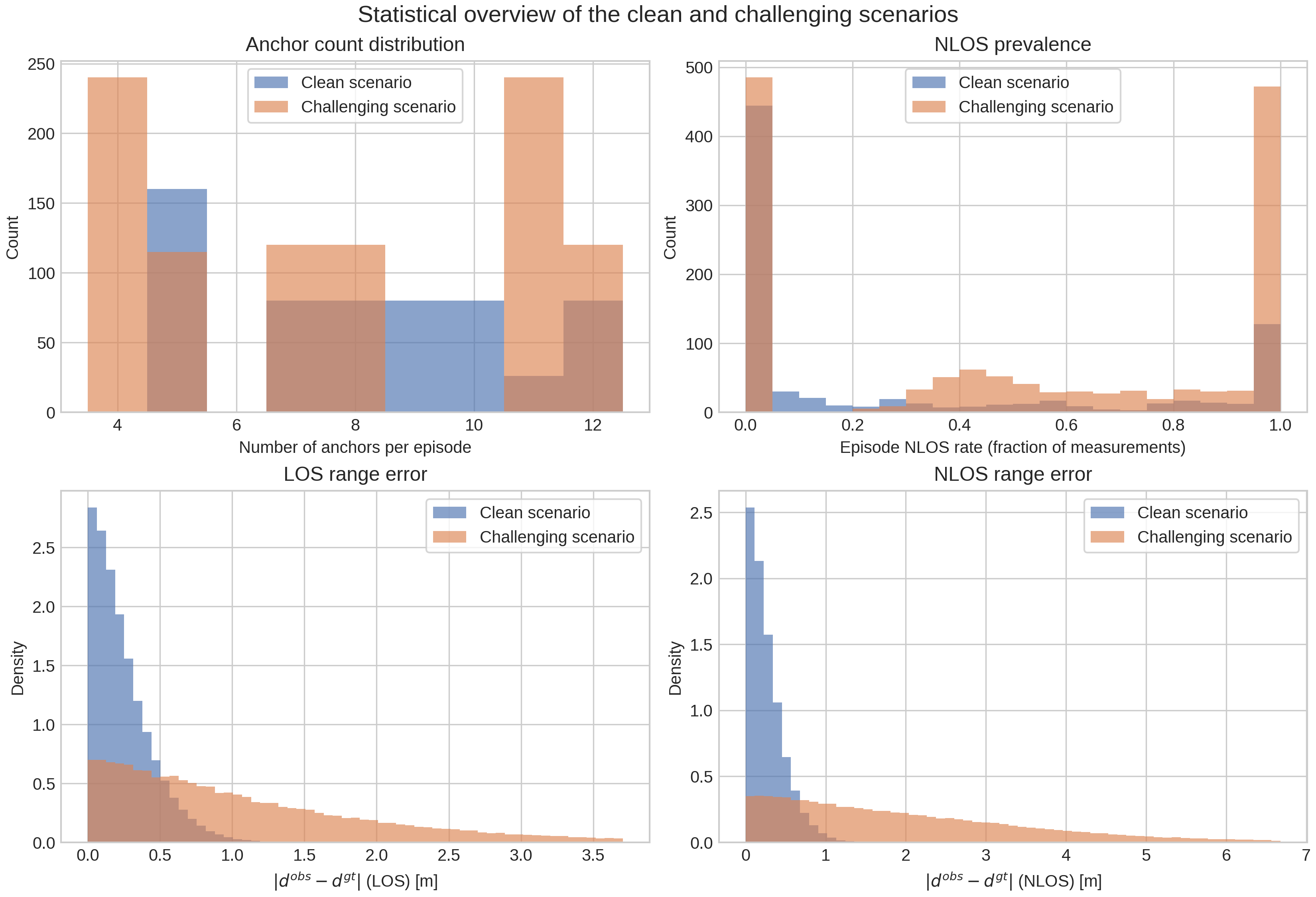}
  \caption{Statistical overview of the clean and challenging scenarios.}
  \label{fig:scenario_overview}
\end{figure}
Figure~\ref{fig:scenario_overview} confirms that the challenging scenario differs from the clean scenario along several axes: it contains more high-NLOS episodes and substantially heavier LOS/NLOS error tails while retaining a comparable anchor-count range. This supports using the challenging scenario as the primary stress-test.

\subsection{Baselines and Comparison Protocol}
\label{sub:baselines}
We compare against four baselines:
\begin{lightdescription}
  \item[Kalman (1D)] independent per-anchor constant-position Kalman filter with scalar process noise $Q$ and measurement noise $R$ tuned on the validation split.
  \item[MLP] fixed-order per-step denoiser that consumes the padded multi-anchor range--mask vector without pose input.
  \item[PoseMLP] fixed-order per-step denoiser that consumes the padded multi-anchor range--mask vector together with pose input.
  \item[PoseKalman] adaptive Rauch--Tung--Striebel (RTS) smoother with pose-dependent process noise and NLOS measurement inflation. The inflation uses the simulation-only indicator $n_{t,i}$, so PoseKalman is an oracle-assisted baseline in the simulation benchmarks. On field data, where no label is available, it runs as the adaptive smoother without NLOS inflation.
\end{lightdescription}

All learned models use a strict episode-level $70/15/15$ train/val/test split to avoid trajectory leakage. Metric masks (overall/LOS/NLOS), stress-test protocols, and downstream solvers are shared across methods, which keeps the comparison focused on the denoising backbone rather than protocol asymmetry. Checkpoint selection is validation-driven for every method and the per-method selection criterion is stated in the implementation details below.

\subsection{Downstream Solvers and Evaluation Metrics}
\label{sec:downstream_eval}
All methods feed their denoised ranges into the same downstream solver, so any downstream difference can be attributed to the input ranges. The deployment-relevant downstream task is cone localization under a known tag trajectory, which avoids the confounding of joint simultaneous localization and mapping (SLAM): the RTK/INS reference provides the tag positions $\{p_t\}$, and the denoised ranges $\hat d_{t,i}$ are used to estimate each cone (anchor) position $\hat a_i$ by a weighted nonlinear least-squares fit:
\begin{equation}
\hat a_i = \arg\min_{a\in\mathbb{R}^2}\ \sum_t w^{map}_{t,i}\big(\|p_t-a\|_2 - \hat d_{t,i}\big)^2,
\end{equation}
solved by Gauss--Newton with a linearized least-squares (LS) initializer and damped refinement. The weights are $w^{map}_{t,i}=m_{t,i}[(1-n_{t,i})+\alpha n_{t,i}]$ with $\alpha=0.2$ when the oracle $n_{t,i}$ is available (simulation only), applied identically to all methods. These oracle-NLOS weights are a simulation-only diagnostic, and a deployment-facing pipeline would replace the term with an estimated reliability weight or simply use $w^{map}_{t,i}=m_{t,i}$. The recovered cone positions are then assembled into the work zone polygon, defined as the convex hull (the smallest convex polygon enclosing the mapped anchors) over the valid-anchor index set of an episode. For episodes that retain fewer than three valid mapped anchors, the convex hull is degenerate. Polygon intersection-over-union (IoU) and Hausdorff distance (the largest boundary-to-boundary gap between two polygons) are undefined for these episodes and are excluded from the reported polygon averages, which are taken over the non-degenerate episodes.

We report range MSE on LOS/NLOS/overall subsets together with tail percentiles ($p50$--$p99$), burst max-error and recovery duration, and anchor permutation/dropout stress-test MSE for signal-level quality. For downstream we report Anchor Error$_w$ (mean Euclidean error over valid anchors, measuring cone-position accuracy) in the main comparison, and polygon IoU plus symmetric Hausdorff distance $d_H$ (measuring work zone geometry) in the model-selection and robustness analyses. Episode-level Spearman correlations between range MSE and downstream metrics, and degradation under injected planar pose noise, complete the diagnostic set. Polygon metrics can be non-monotonic in range MSE because the convex-hull proxy is sensitive to extreme anchors, which is why we report both signal-level and downstream metrics throughout.

\subsection{Implementation Details and Model Selection}

The final configuration of the proposed method uses hidden size $256$, decoder hidden size $256$, mean pooling, and context-based decoding. Stage 1, the model is pretrained on observed targets. Stage 2 then fine-tunes the model against ground-truth ranges with the weighted Huber loss on the full labeled training split. In the retained challenging-scenario checkpoint we set $\lambda_{pred}=0$ during Stage 2, so fine-tuning optimizes only the supervised denoising loss once predictive pretraining has initialized the backbone.

The learned baselines (MLP and PoseMLP) retain the checkpoint with the lowest validation range loss. For the challenging scenario, the proposed method is additionally selected on the validation split by downstream polygon IoU under the fixed solver in Section~\ref{sec:downstream_eval}. This polygon-tuned checkpoint is then reused throughout the main, robustness, and downstream analyses, and the checkpoint-selection asymmetry relative to the base checkpoint is analyzed in Table~\ref{tab:polygon_tuned}. The field evaluation uses the matched baseline implementations and reports the additional proposed-denoiser output calibration explicitly in the results section. All simulation results below are computed on held-out test data.

\section{Results}

Results follow the evaluation protocol in Section~\ref{sec:exp_setup}. The field study first tests whether the pipeline can be applied to measured V2I UWB data. The controlled simulation study then isolates the denoising contribution, downstream effects, anchor-set robustness, and deployment sensitivities under known ground truth.

\subsection{Field Feasibility on Measured V2I Data}

The field track has eight dynamic driving episodes with varying anchor visibility and two static reference logs. The preprocessing pipeline of Section~\ref{sec:exp_setup} converts asynchronous UWB events into aligned pose--range--mask sequences, so the simulation-trained denoisers and shared downstream solvers apply directly without interface changes. All methods use the same per-anchor affine input calibration. The proposed denoiser is also reported with a degree-2 output calibration on its field outputs. Because this calibration uses field reference information, the corresponding result should be interpreted as a calibrated field diagnostic.

\paragraph{Signal-level range accuracy.}
The signal-level columns of Table~\ref{tab:real_results} report range error on the dynamic field episodes, weighted by valid measurement count across all eight episodes. With the shared input calibration and the additional degree-2 output calibration described above, the proposed denoiser obtains the lowest MSE ($1.418$\,m$^2$) and MAE ($0.750$\,m). This corresponds to a $66.9\%$ MSE and $50.8\%$ MAE reduction relative to the calibrated raw input ($4.286$\,m$^2$ MSE, $1.524$\,m MAE). PoseKalman is the closest baseline at $4.261$\,m$^2$ MSE and $1.536$\,m MAE, essentially at parity with the calibrated raw input. MLP and PoseMLP yield higher errors than the calibrated raw input, indicating that the simulation-trained feed-forward baselines do not match this field distribution under the calibration used here. Because the proposed denoiser includes an output calibration fitted with field reference information, the field numbers should be read as calibrated deployment diagnostics.

\paragraph{Downstream localization and mapping.}
The downstream columns of Table~\ref{tab:real_results} report the corresponding geometry metrics, using the same solver as in simulation. The absolute downstream scale is set by the small site footprint and by the fact that the reference distances come from measured site dimensions plus RTK/INS rather than per-frame surveyed truth. Within this regime, the calibrated output of the proposed denoiser gives the lowest Anchor Error$_w$ ($12.23$\,m) and the highest polygon IoU ($0.046$), the two cone-layout metrics on which it ranks first among the compared field pipelines.

\begin{table*}[ht]
\centering
\caption{Real-world dynamic-episode results. Signal-level range error is weighted by valid measurement count. Downstream geometry uses the shared solvers. The proposed field row includes the degree-2 output calibration fitted to field reference distances described in the text. Lower is better except IoU. Best per column in bold, second best underlined.}
\label{tab:real_results}
\begin{tabular}{lccccc}
\toprule
 & \multicolumn{2}{c}{\textbf{Signal-level range error}} & \multicolumn{3}{c}{\textbf{Downstream geometry}} \\
\cmidrule(lr){2-3} \cmidrule(lr){4-6}
\textbf{Method} & \textbf{MSE $\downarrow$} & \textbf{MAE $\downarrow$} & \textbf{Anchor Error$_w$ $\downarrow$} & \textbf{IoU $\uparrow$} & \textbf{Hausdorff $\downarrow$} \\
\midrule
Raw (calibrated input)   & 4.286  & \underline{1.524} & \underline{13.631} & 0.000 & \textbf{12.395} \\
Kalman (1D)              & 12.478 & 2.533 & 38.827 & \underline{0.045} & 56.029 \\
MLP                      & 13.596 & 3.101 & 18.088 & 0.013 & 29.589 \\
PoseMLP                  & 26.364 & 4.638 & 27.062 & 0.015 & 71.999 \\
PoseKalman               & \underline{4.261}  & 1.536 & 13.989 & 0.000 & \underline{14.372} \\
\textbf{Proposed method (field-calibrated)} & \textbf{1.418} & \textbf{0.750} & \textbf{12.225} & \textbf{0.046} & 22.858 \\
\bottomrule
\end{tabular}
\end{table*}

\begin{figure}[]
  \centering
  \includegraphics[width=\linewidth]{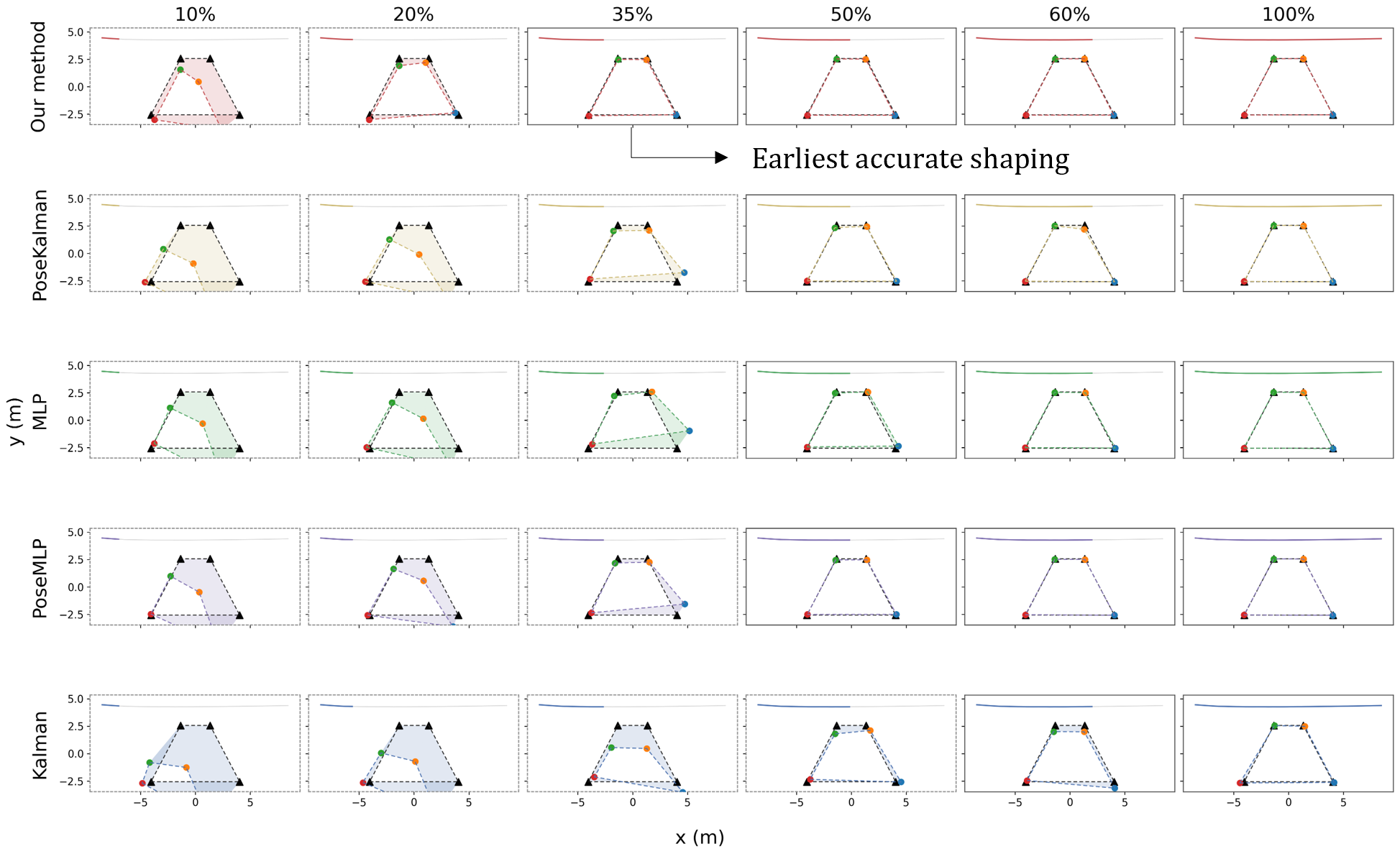}
  \caption{Progressive work zone geometry reconstruction on one real dynamic episode. Rows: methods. Columns: fraction of the episode used ($10$--$100\%$). Black triangles: geometry-derived reference trapezoidal proxy. Colored markers: estimated anchor positions. The annotation marks the earliest panel where the proposed method recovers a recognizable trapezoid.}
  \label{fig:real_visualization}
\end{figure}

\paragraph{Representative progressive work zone geometry reconstruction.}
Figure~\ref{fig:real_visualization} compares the reconstructed anchor layouts as the vehicle accumulates observations across one dynamic episode ($10$--$100\%$ of the episode). The proposed method forms a recognizable trapezoid by $20\%$ and stabilizes by $35\%$. PoseKalman and MLP require more observations before approaching the correct geometry, while PoseMLP and Kalman are less stable in the early fractions and remain less competitive in the aggregate downstream metrics. This representative episode illustrates the qualitative behavior behind the aggregate anchor-error and polygon-IoU results.

\subsection{Range Denoising in Controlled Simulation}

The simulation study uses the challenging and clean regimes of Section~\ref{sec:data_gen} to isolate denoising performance under controlled conditions. The challenging regime is the target stress case for bursty NLOS and heavy-tailed residuals, whereas the clean regime serves as a negative control showing when simple filtering remains preferable.

\subsubsection{Range Denoising Across Challenging and Clean Regimes}
\begin{table*}[ht]
\centering
\caption{Range denoising MSE on the challenging and clean test splits (lower is better. Best per column in bold, second best underlined).}
\label{tab:sim_mse}
\begin{tabular}{lcccccc}
\toprule
 & \multicolumn{3}{c}{\textbf{Challenging scenario}} & \multicolumn{3}{c}{\textbf{Clean scenario}} \\
\cmidrule(lr){2-4} \cmidrule(lr){5-7}
\textbf{Method} & \textbf{MSE$_{LOS}$ $\downarrow$} & \textbf{MSE$_{NLOS}$ $\downarrow$} & \textbf{MSE$_{Overall}$ $\downarrow$} & \textbf{MSE$_{LOS}$ $\downarrow$} & \textbf{MSE$_{NLOS}$ $\downarrow$} & \textbf{MSE$_{Overall}$ $\downarrow$} \\
\midrule
Kalman (1D)              & 1.6736 & 4.7976 & 4.1165 & \underline{0.0360} & \textbf{0.0340} & \textbf{0.0353} \\
MLP                      & 1.3002 & 3.4651 & 2.9931 & 0.2493 & 0.3207 & 0.2755 \\
PoseMLP                  & 0.7834 & 2.6400 & 2.2352 & 0.0653 & 0.1416 & \underline{0.0933} \\
PoseKalman               & \underline{0.5636} & \underline{2.0814} & \underline{1.7505} & \textbf{0.0313} & 0.5179 & 0.2097 \\
\textbf{Proposed method} & \textbf{0.5285} & \textbf{0.7075} & \textbf{0.6686} & 0.1042 & \underline{0.1247} & 0.1117 \\
\bottomrule
\end{tabular}
\end{table*}

The pattern is regime-dependent. In the challenging scenario, the proposed method reduces overall MSE by $77.7\%$ relative to MLP, $83.8\%$ relative to Kalman, and $61.8\%$ relative to PoseKalman, with NLOS MSE $66.0\%$ lower than PoseKalman. In the clean scenario, Kalman remains optimal on overall MSE. But the proposed method still shows competitive performance. This indicates that the proposed architecture is most useful when non-Gaussian NLOS errors and temporal range corruption dominate the measurement process, which appear frequently in the real traffic scenarios. 

\begin{figure}[]
  \centering
  \includegraphics[width=0.98\linewidth]{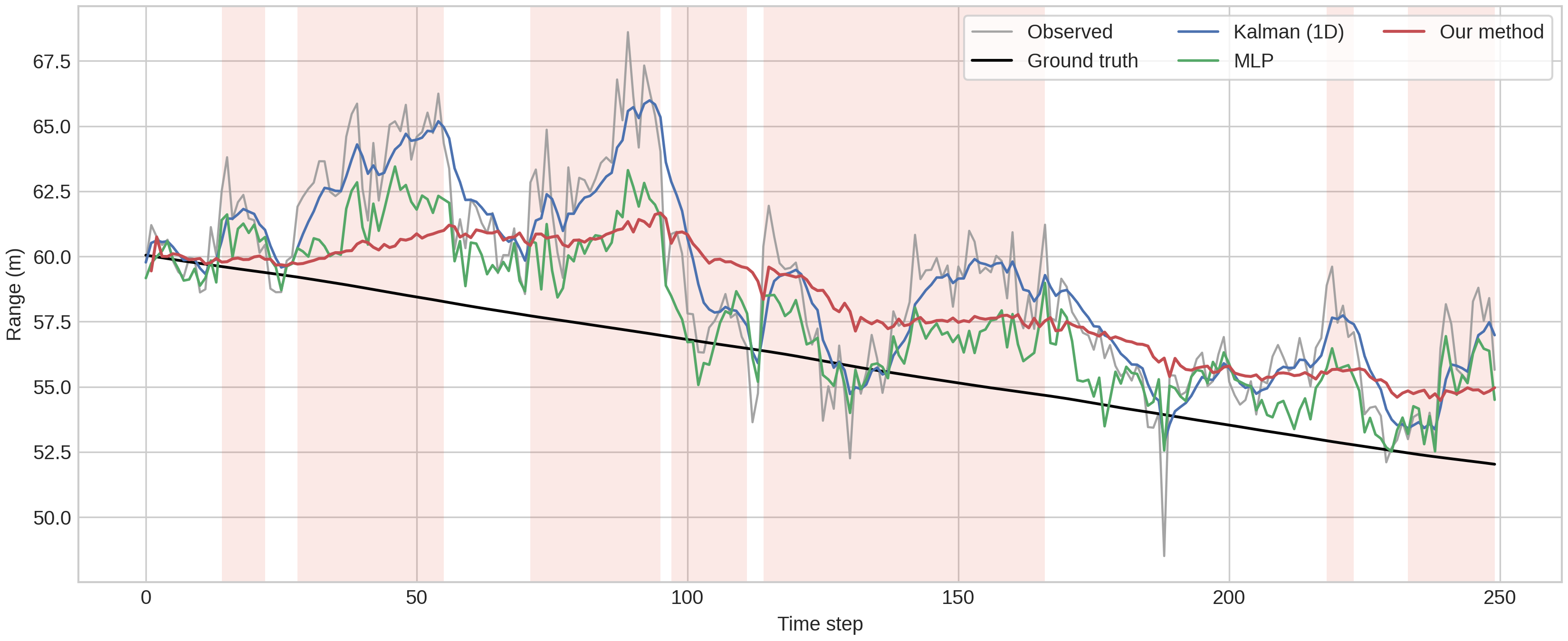}
  \caption{Time-series comparison on one anchor stream from the challenging scenario. The shaded regions indicate NLOS-active segments.}
  \label{fig:distance_vs_time_challenging}
\end{figure}
Figure~\ref{fig:distance_vs_time_challenging} gives a sequence-level view of the same effect: across extended NLOS segments, the proposed method suppresses the large positive excursions seen in the raw observations, MLP, and Kalman, and follows the downward ground-truth trend more stably, although residual positive bias remains in some long NLOS intervals.

\subsubsection{Polygon-Oriented Model Selection}
The pipeline is ultimately judged by downstream geometry rather than by range loss alone, so we select the retained checkpoint on the validation split using polygon IoU under the fixed downstream solver. The challenging-scenario row for the proposed method in Table~\ref{tab:sim_mse} already uses this polygon-tuned checkpoint. Table~\ref{tab:polygon_tuned} compares it with the base checkpoint of the same architecture.
\begin{table}[ht]
\centering
\small
\caption{Base vs. polygon-tuned checkpoints of the proposed method on the challenging scenario test split.}
\label{tab:polygon_tuned}
\begin{tabular}{lccc}
\toprule
\textbf{Variant} & \textbf{MSE$_{Overall}$ $\downarrow$} & \textbf{IoU $\uparrow$} & \textbf{Hausdorff $\downarrow$} \\
\midrule
Proposed method (base checkpoint)          & 0.789 & 0.297 & 56.65 \\
Proposed method (polygon-tuned checkpoint) & \textbf{0.669} & \textbf{0.315} & \textbf{53.78} \\
\bottomrule
\end{tabular}
\end{table}
The polygon-tuned checkpoint improves the base checkpoint by $15.2\%$ in overall MSE and about $6\%$ in polygon IoU. We therefore use this checkpoint for the remaining challenging-regime analyses. This model-selection rule is used because range MSE is only partially aligned with polygon quality. For the proposed method, the episode-level Spearman correlations of range MSE with polygon IoU and Hausdorff are $-0.320$ and $0.362$, respectively.

\subsubsection{Statistical Robustness of Main Results}
A single-point MSE can be misleading on a finite test set, so we report episode-level bootstrap confidence intervals (1000 resamples). Table~\ref{tab:bootstrap_main} shows the mean overall MSE and the $95\%$ CI for both benchmark regimes. These episode-weighted means need not match the measurement-weighted MSEs of Table~\ref{tab:sim_mse} exactly, because episodes contain different numbers of valid measurements.
\begin{table}[ht]
\centering
\small
\caption{Bootstrap robustness for episode-weighted overall MSE (mean of episode-level MSEs [95\% CI]). Best per column in bold.}
\label{tab:bootstrap_main}
\begin{tabular}{lcc}
\toprule
\textbf{Method} & \textbf{challenging scenario $\downarrow$} & \textbf{clean scenario $\downarrow$} \\
\midrule
Kalman     & 4.173 [4.039, 4.306] & \textbf{0.0349} [0.0336, 0.0362] \\
MLP        & 3.133 [2.957, 3.304] & 0.276 [0.260, 0.292] \\
PoseMLP    & 2.471 [2.218, 2.745] & 0.0916 [0.0845, 0.0989] \\
PoseKalman & 1.760 [1.650, 1.867] & 0.191 [0.152, 0.230] \\
\textbf{Proposed method} & \textbf{0.665} [0.627, 0.702] & 0.111 [0.107, 0.115] \\
\bottomrule
\end{tabular}
\end{table}
The challenging-regime confidence interval for the proposed method does not overlap with any baseline, indicating that the main MSE improvement is statistically stable.

\subsection{Cone Localization, Work Zone Geometry, and Anchor-Set Robustness}
\label{sec:robust}
This subsection covers cone-position accuracy on the challenging-scenario benchmark and the robustness of cone localization and work zone geometry reconstruction under deployment-relevant anchor perturbations. We first report range-level tail, burst, and stress behavior, then evaluate downstream cone-position and polygon-shape stability under episode-consistent anchor re-indexing and anchor dropout.

\subsubsection{Tail, Burst, and Stress Robustness}
\begin{figure}[]
  \centering
  \includegraphics[width=0.98\linewidth]{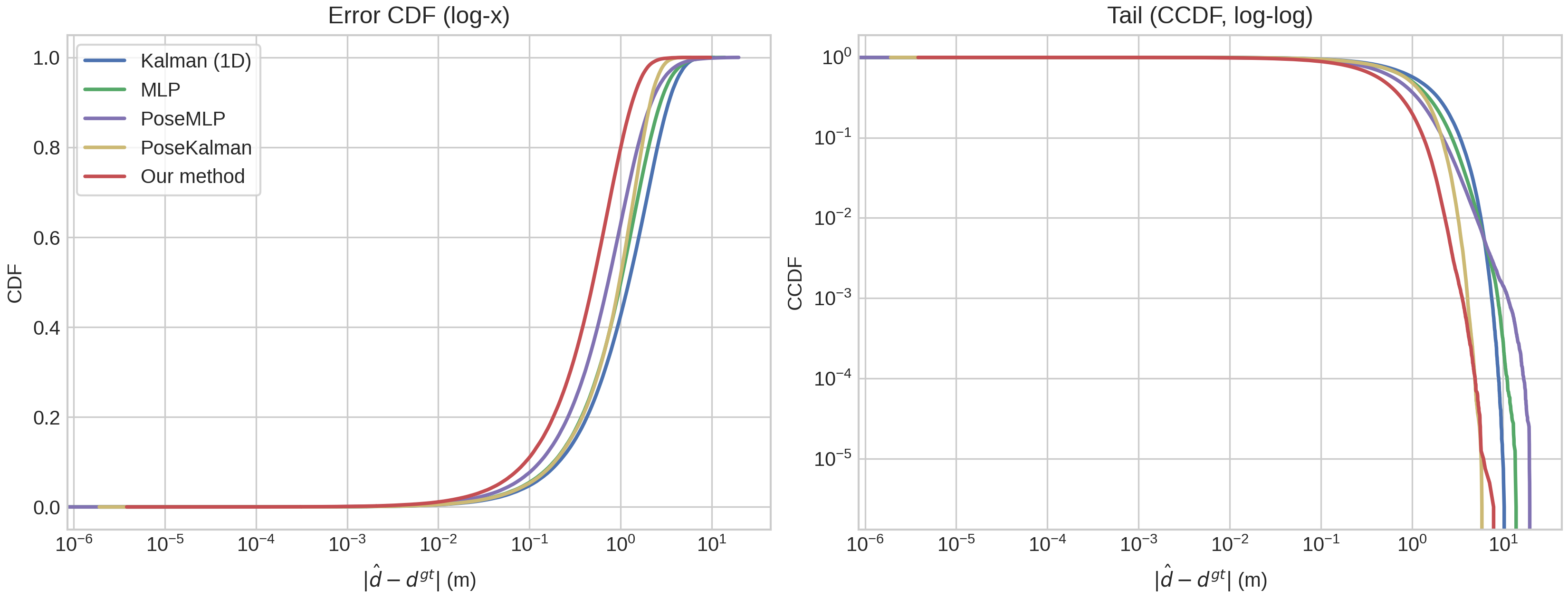}
  \caption{Absolute-error CDF (left) and complementary CDF on log-log axes (right) in the challenging scenario. The proposed method shifts the distribution leftward and has the lightest extreme-error tail.}
  \label{fig:tail_error_cdf_ccdf}
\end{figure}

Table~\ref{tab:tail_burst} shows that the proposed method has the lightest error tail and the lowest burst max error.  Figure~\ref{fig:tail_error_cdf_ccdf} plots the corresponding error distribution. Recovery duration is not uniformly minimized, suggesting a trade-off between stable burst suppression and rapid transient correction.
\begin{table}[ht]
\centering
\small
\caption{Tail percentiles (overall absolute error, m) and burst-segment metrics on the challenging scenario test split. Lower is better. Best per column in bold.}
\label{tab:tail_burst}
\begin{tabular}{lccccc}
\toprule
\textbf{Method} & \textbf{Tail $p95$ $\downarrow$} & \textbf{Tail $p99$ $\downarrow$} & \textbf{Burst max mean $\downarrow$} & \textbf{Burst max $p95$ $\downarrow$} & \textbf{Recovery $p95$ $\downarrow$} \\
\midrule
Kalman      & 4.140 & 5.695 & 3.576 & 6.847 & 20 \\
MLP         & 3.453 & 5.339 & 3.075 & 6.114 & 10 \\
PoseMLP     & 2.893 & 5.095 & 2.229 & 5.366 & \textbf{6} \\
PoseKalman  & 2.480 & 3.205 & 1.288 & 3.185 & 41 \\
\textbf{Proposed method} & \textbf{1.644} & \textbf{2.302} & \textbf{1.152} & \textbf{2.574} & 17 \\
\bottomrule
\end{tabular}
\end{table}

\begin{table}[ht]
\centering
\small
\caption{Anchor-stress robustness in the challenging scenario: overall MSE under anchor dropout rates from $0.0$ to $0.5$ (mean MSE), and under $20$ random episode-consistent permutations (mean $\pm$ std across permutations). The Drop $0.0$ column is the original stress-test reference and is approximately the same as the nominal challenging-scenario MSE in Table~\ref{tab:sim_mse}, with small differences due to rounding and stress-test evaluation details. The best result in each column is shown in bold.}
\label{tab:stress}
\begin{tabular}{lccccc}
\toprule
\textbf{Method} & \textbf{Drop 0.0 $\downarrow$} & \textbf{Drop 0.1 $\downarrow$} & \textbf{Drop 0.3 $\downarrow$} & \textbf{Drop 0.5 $\downarrow$} & \textbf{Perm MSE $\downarrow$} \\
\midrule
Kalman      & 4.128   & 4.117   & 4.143   & 4.182   & $4.128\pm0.000$  \\
MLP         & 3.002   & 34.942  & 96.539  & 193.575 & $149.909\pm5.062$ \\
PoseMLP     & 2.242   & 32.633  & 127.643 & 299.089 & $189.840\pm6.282$ \\
PoseKalman  & 1.743   & 1.742   & 1.723   & 1.797   & $1.743\pm0.000$  \\
\textbf{Proposed method} & \textbf{0.669} & \textbf{0.664} & \textbf{0.687} & \textbf{0.705} & $\mathbf{0.669\pm0.000}$ \\
\bottomrule
\end{tabular}
\end{table}

\begin{figure}[]
  \centering
  \includegraphics[width=0.60\linewidth]{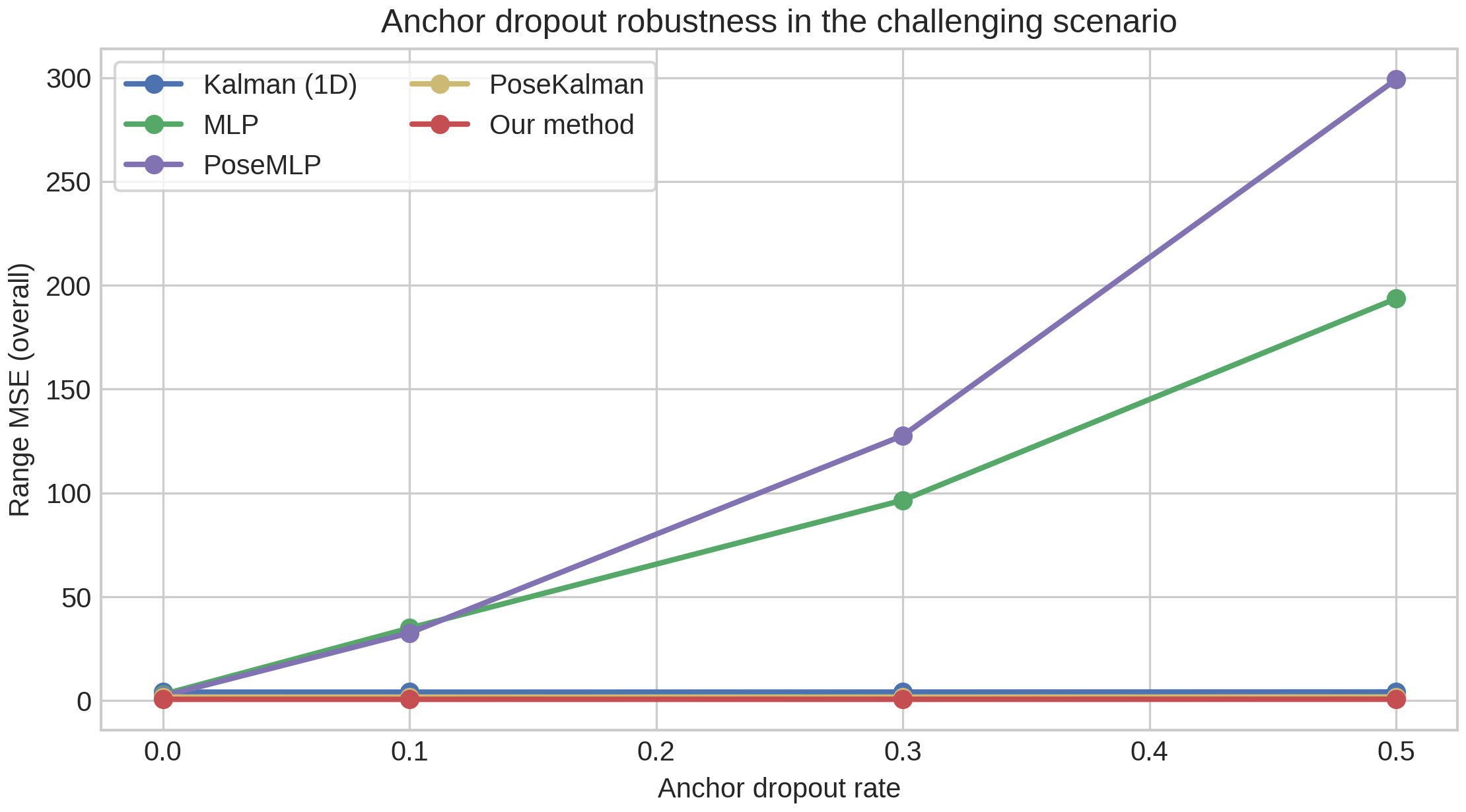}
  \caption{Anchor dropout robustness on the challenging scenario.}
  \label{fig:dropout_robustness}
\end{figure}

In Table~\ref{tab:stress}, the Drop $0.0$ column applies no perturbation which means the full anchor set is in its canonical order. And therefore it serves as the unperturbed reference corresponding to the nominal challenging-scenario MSE. The remaining columns apply either anchor dropout or anchor re-indexing. The proposed method is insensitive to episode-consistent anchor permutations and remains stable under the anchor-dropout sweep: range MSE stays at $0.669$ across all $20$ permutations and remains below the baseline values at dropout rates up to $0.5$. PoseKalman is also invariant to permutation, and both it and Kalman stay flat under dropout, because all three of these methods process anchors independently or with order-symmetric operators. In contrast, MLP and PoseMLP consume the multi-anchor observation as a fixed-order vector, so episode-consistent re-indexing reshuffles their input and the learned per-slot mapping no longer applies. Their permutation MSE rises to $150$--$190$, and both degrade sharply under dropout as well ($193.6$ and $299.1$ at $50\%$). PoseMLP is the more brittle of the two despite its lower nominal MSE: conditioning on pose lets it fit the canonical anchor ordering more tightly, which makes it degrade further once that ordering is broken. Figure~\ref{fig:dropout_robustness} shows the corresponding dropout curves.

\subsubsection{Cone Localization and Work Zone Shape Under Anchor Perturbations}
All downstream metrics use the fixed localization pipeline of Section~\ref{sec:downstream_eval} with only the input ranges differing across methods. The proposed-method row uses the unified polygon-tuned checkpoint retained above, and baseline rows use their matched retained checkpoints.
\begin{table}[ht]
\centering
\caption{Cone-position accuracy (Anchor Error$_w$, under canonical anchor ordering) and work zone shape stability (polygon IoU, mean $\pm$ std over 20 random episode-consistent anchor permutations) on the challenging scenario test split. Lower is better except IoU. Best per column in bold, second best underlined.}
\label{tab:downstream}
\small
\begin{tabular}{lccc}
\toprule
\textbf{Method} & \textbf{Anchor Error$_w$ $\downarrow$} & \textbf{IoU $\uparrow$} & \textbf{IoU std $\downarrow$} \\
\midrule
Kalman     & 33.16 & 0.264 & 0.000 \\
MLP        & 32.07 & 0.221 & 0.007 \\
PoseMLP    & \textbf{27.83} & 0.218 & 0.006 \\
PoseKalman & 35.05 & \underline{0.268} & 0.000 \\
\textbf{Proposed method} & \underline{29.25} & \textbf{0.313} & 0.001 \\
\bottomrule
\end{tabular}
\end{table}

\begin{table}[ht]
\centering
\caption{Downstream robustness under anchor dropout on the challenging scenario. Polygon IoU is reported for selected dropout rates. Best per column in bold. Polygon IoU is averaged only over non-degenerate episodes with at least three retained mapped anchors.}
\label{tab:dropout_downstream}
\small
\begin{tabular}{lccc}
\toprule
\textbf{Method} & \textbf{IoU@0.0 $\uparrow$} & \textbf{IoU@0.3 $\uparrow$} & \textbf{IoU@0.5 $\uparrow$} \\
\midrule
MLP        & 0.284 & 0.116 & 0.052 \\
PoseKalman & 0.268 & 0.213 & 0.172 \\
\textbf{Proposed method} & \textbf{0.315} & \textbf{0.236} & \textbf{0.230} \\
\bottomrule
\end{tabular}
\end{table}

For compactness, Table~\ref{tab:dropout_downstream} reports a representative subset: the index-sensitive MLP, the strongest invariant baseline PoseKalman, and the proposed method.

Table~\ref{tab:downstream} reports cone-position accuracy under the canonical anchor ordering (Anchor Error$_w$) alongside work zone shape stability under permutation (polygon IoU mean $\pm$ std over 20 random re-indexings). PoseMLP has the lowest canonical Anchor Error$_w$ at $27.83$\,m, with the proposed method second at $29.25$\,m, but this canonical-order advantage is not robust to re-indexing: PoseMLP's polygon IoU under permutation drops to $0.218$ with non-negligible std ($0.006$), whereas the proposed method preserves polygon IoU at $0.313$ with std $0.001$ and ranks first on the work zone shape metric. PoseKalman is similarly invariant. Anchor dropout gives a more nuanced picture (Table~\ref{tab:dropout_downstream}): at $30\%$ dropout the proposed method has the highest polygon IoU ($0.236$) among the reported methods, but at $50\%$ dropout the polygon quality degrades for every method even though range MSE for the equivariant denoisers stays flat. This is observability-limited. $28.2\%$ of episodes retain only two anchors after $50\%$ masking and become weakly constrained for cone localization (recovering an anchor position from range measurements, which needs several well-spread tag observations). And even here the proposed method retains the highest polygon IoU ($0.230$ vs $0.172$ for PoseKalman and $0.052$ for MLP).
\begin{figure}[]
  \centering
  \includegraphics[width=0.98\linewidth]{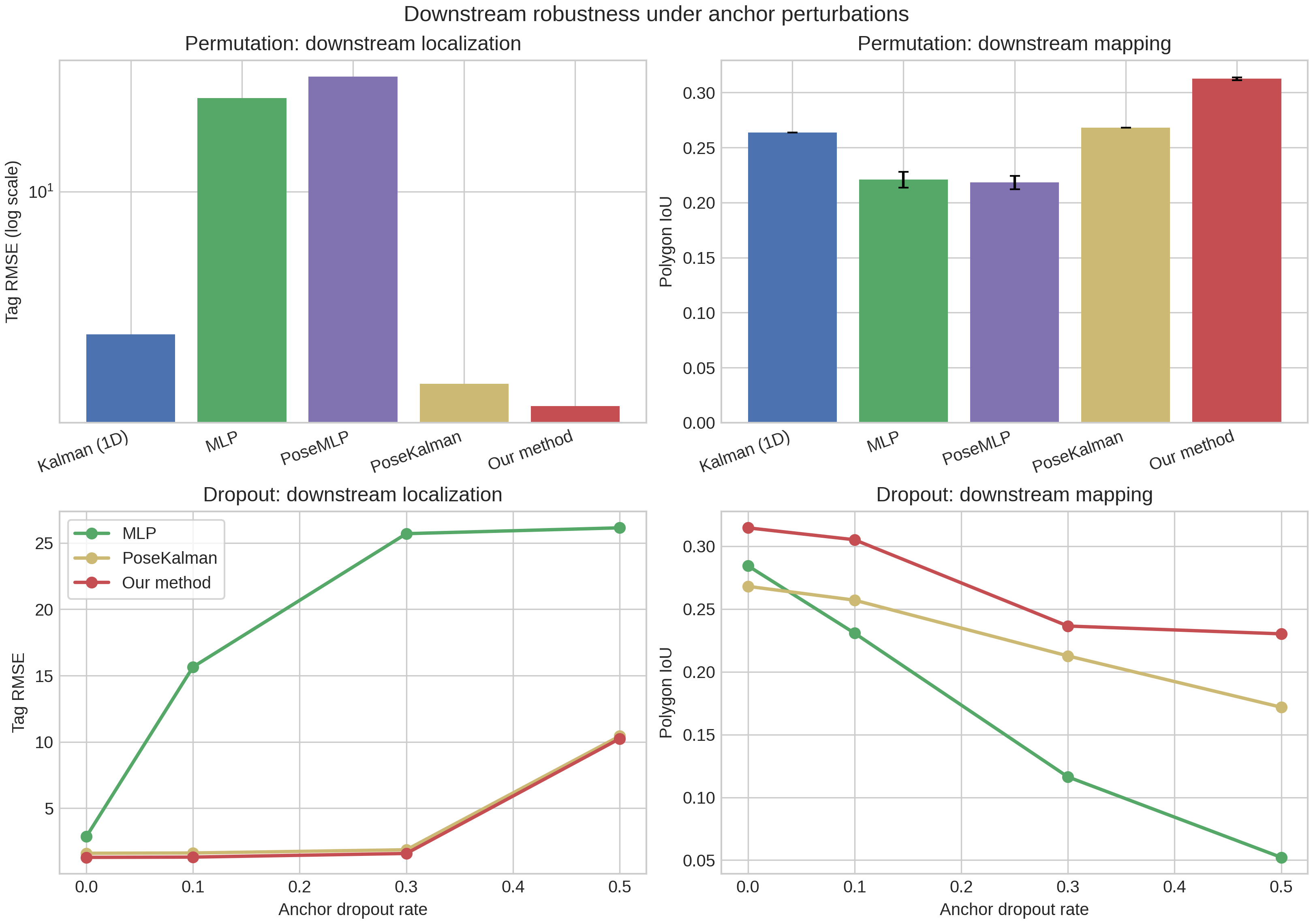}
  \caption{Downstream robustness under anchor perturbations on the challenging scenario. Top: permutation results for all compared methods. Bottom: dropout results for the representative subset reported in Table~\ref{tab:dropout_downstream}.}
  \label{fig:downstream_robustness_summary}
\end{figure}
Figure~\ref{fig:downstream_robustness_summary} visualizes both regimes side by side.

\subsection{Component and Deployment Sensitivity}
The final analyses examine how strongly the retained model depends on its architectural information pathways and on the quality of the pose input. These tests should be interpreted as component and deployment sensitivities rather than as additional primary benchmarks.

\subsubsection{Component Ablation}
We measure the contribution of each design choice by ablating it on the unified challenging-scenario checkpoint: pretraining removed before fine-tuning, set context removed at runtime, and pose input removed at runtime. The context and pose rows are runtime-removal sensitivities, so they indicate how strongly the trained model relies on each information pathway rather than serving as fully retrained architectural ablations. Table~\ref{tab:ablation_stagea} reports the resulting metrics.
\begin{table}[ht]
\centering
\small
\caption{Component ablation on the challenging scenario test split (family of the proposed method). Best per column in bold.}
\label{tab:ablation_stagea}
\begin{tabular}{lccc}
\toprule
\textbf{Variant} & \textbf{MSE$_{Overall}$ $\downarrow$} & \textbf{MSE$_{NLOS}$ $\downarrow$} & \textbf{IoU $\uparrow$} \\
\midrule
Full model            & \textbf{0.669} & \textbf{0.707} & \textbf{0.315} \\
w/o pretraining       & 0.937 & 1.008 & 0.286 \\
w/o context (runtime) & 1.079 & 1.064 & 0.244 \\
w/o pose (runtime)    & 6.810 & 7.741 & 0.131 \\
\bottomrule
\end{tabular}
\end{table}
Pose conditioning is the largest observed dependency in this runtime-removal sensitivity test: removing it at runtime raises overall MSE by an order of magnitude (from $0.669$ to $6.810$) and substantially degrades the trained model relative to the other tested variants. Set context is the next largest observed dependency, followed by predictive pretraining. This ordering holds for both overall MSE and polygon IoU. The ``w/o pretraining'' row also quantifies the design role of Stage 1 described in Section~\ref{sec:peq_denoiser}: omitting Stage 1 and training the same backbone from random initialization under supervised loss raises overall MSE by $40.1\%$ and NLOS MSE by $42.6\%$, with the largest benefit in the NLOS-heavy regime where the supervised loss alone is most non-convex.

\subsubsection{Pose-Noise Sensitivity}
\begin{table}[ht]
\centering
\caption{Pose-noise sensitivity on the challenging scenario when zero-mean planar noise is injected only into the model pose input at test time. Best per column in bold.}
\label{tab:pose_noise}
\small
\begin{tabular}{lccc}
\toprule
\textbf{Method} & \textbf{MSE@0.0 $\downarrow$} & \textbf{MSE@0.25 $\downarrow$} & \textbf{MSE@0.5 $\downarrow$} \\
\midrule
PoseMLP    & 2.242 & 2.250 & 2.274 \\
PoseKalman & 1.743 & 2.739 & 3.415 \\
\textbf{Proposed method} & \textbf{0.669} & \textbf{0.727} & \textbf{0.912} \\
\bottomrule
\end{tabular}
\end{table}

\begin{figure}[]
  \centering
  \includegraphics[width=0.98\linewidth]{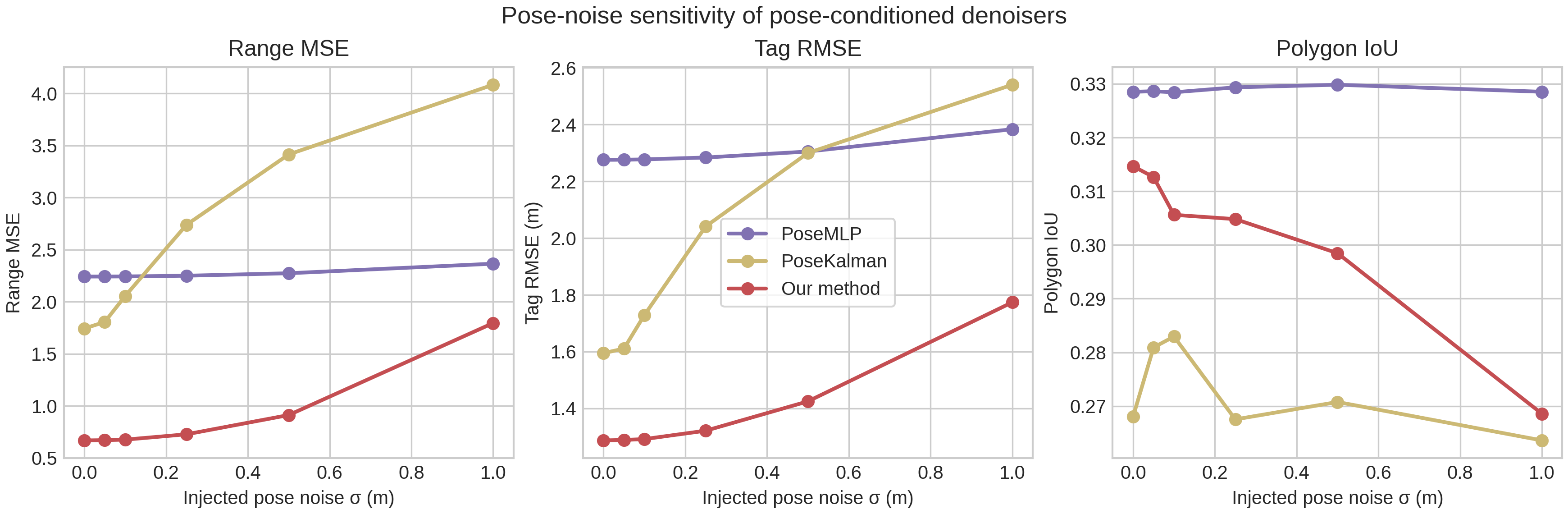}
  \caption{Pose-noise sensitivity of pose-conditioned denoisers on the challenging scenario.}
  \label{fig:pose_noise_sensitivity}
\end{figure}

Table~\ref{tab:pose_noise} isolates the dependence on pose quality. It lists only the pose-conditioned methods, since pose-free Kalman and MLP are unaffected by perturbations applied to the pose input. For range MSE, the proposed method stays the most accurate across the injected-noise levels in the table: overall MSE rises from $0.669$ to $0.727$ at $0.25$\,m and to $0.912$ at $0.50$\,m. PoseMLP is flatter but markedly less accurate, while PoseKalman is more sensitive to corrupted pose. Polygon IoU is only partially aligned with range denoising: the proposed method's IoU decreases with pose noise, from $0.315$ to $0.298$ at $0.50$\,m. The larger $1.0$\,m perturbation point is shown in Figure~\ref{fig:pose_noise_sensitivity} and omitted from the compact table. Moderate pose error therefore does not invert the range MSE ranking among these methods.
Figure~\ref{fig:pose_noise_sensitivity} shows the curves.

\section{Conclusion}

This paper framed work zone mapping with UWB-RSUs as a V2I-enabled geometry-inference problem in which multi-anchor range quality directly affects cone localization and work zone geometry reconstruction. We proposed a pose-conditioned, permutation-equivariant predictive denoiser that combines shared temporal modeling, symmetric set aggregation, pose-conditioned residual decoding, and a two-stage training protocol.

The results support three main conclusions. First, in the challenging simulation regime with bursty NLOS, heavy-tailed outliers, and anchor-set perturbations, the proposed method gives the best range-denoising and work zone geometry reconstruction performance among the evaluated baselines. Second, the robustness analyses show that permutation equivariance is not only a model property at the signal level but also carries into downstream geometry under anchor re-indexing, while severe anchor dropout eventually becomes limited by multilateration observability. Third, the real V2I field study shows that the same pipeline can be applied to measured UWB-RSU data and, when reported with the stated field output calibration, produces the most accurate calibrated signal-level ranges and the best field anchor-error and polygon-IoU results among the compared methods.

\section{Acknowledgments}
This study is supported by the Center for Connected and Automated Transportation (CCAT), University of Michigan, through Grant No. 69A3552348305. The views expressed are those of the authors.

\bibliographystyle{cas-model2-names}
\bibliography{ref}

@article{dehman2021workzone_cav_review,
  title   = {Are work zones and connected automated vehicles ready for a harmonious coexistence? A scoping review and research agenda},
  author  = {Dehman, A. and Farooq, B.},
  journal = {Transportation Research Part C: Emerging Technologies},
  volume  = {133},
  pages   = {103422},
  year    = {2021},
  doi     = {10.1016/j.trc.2021.103422}
}

@misc{usdot2024wzdx,
  title        = {Work Zone Data Exchange (WZDx)},
  author       = {{U.S. Department of Transportation}},
  howpublished = {\url{https://www.transportation.gov/av/data/wzdx}},
  year         = {2024},
  note         = {Last updated: 2024-04-30},
  urldate      = {2026-02-07}
}

@inproceedings{seo2022ttcd_crc,
  title     = {Temporary Traffic Control Device Detection for Road Construction Projects using Deep Learning Application},
  author    = {Seo, Sungchul and Chen, Donghui and Kim, Kwangcheol and Kang, Kyubyung and Koo, Dan and Chae, Myungjin and Park, Hyung Keun},
  booktitle = {Construction Research Congress 2022},
  pages     = {392--401},
  year      = {2022},
  publisher = {American Society of Civil Engineers (ASCE)},
  doi       = {10.1061/9780784483961.042}
}

@inproceedings{zuo2023urban_workzone_detection_sizing,
  title     = {Urban Work Zone Detection and Sizing: A Data-Centric Training and Topology-Based Inference Approach},
  author    = {Zuo, Fan and Gao, Jianbo and Ozbay, Kaan and Bian, Zia and Zhang, Zhiqiang},
  booktitle = {2023 IEEE 26th International Conference on Intelligent Transportation Systems (ITSC)},
  pages     = {3235--3240},
  year      = {2023},
  publisher = {IEEE},
  doi       = {10.1109/ITSC57777.2023.10422546}
}

@techreport{habib2018lidar_lane_width_workzones,
  title       = {LiDAR-Based Mobile Mapping System for Lane Width Estimation in Work Zones},
  author      = {Habib, Ayman and Lin, Yun-Jou and Ravi, Radhika and Shamseldin, Tamer and Elbahnasawy, Magdy},
  institution = {Joint Transportation Research Program, Purdue University},
  number      = {FHWA/IN/JTRP-2018/10},
  year        = {2018},
  doi         = {10.5703/1288284316730},
  note        = {SPR-4126}
}

@article{chen2023crowdsourcing_workzone_mapping,
  title         = {Improving Autonomous Vehicle Mapping and Navigation in Work Zones Using Crowdsourcing Vehicle Trajectories},
  author        = {Chen, Hanlin and Luo, Renyuan and Feng, Yiheng},
  journal       = {arXiv preprint arXiv:2301.09194},
  year          = {2023},
  doi           = {10.48550/arXiv.2301.09194},
  eprint        = {2301.09194},
  archivePrefix = {arXiv},
  primaryClass  = {cs.RO}
}

@article{gezici2005uwb_localization_spm,
  title   = {Localization via ultra-wideband radios: a look at positioning aspects for future sensor networks},
  author  = {Gezici, Sinan and Tian, Zhi and Giannakis, Georgios B. and Kobayashi, Hisashi and Molisch, Andreas F. and Poor, H. Vincent and Sahinoglu, Zafer},
  journal = {IEEE Signal Processing Magazine},
  volume  = {22},
  number  = {4},
  pages   = {70--84},
  year    = {2005},
  doi     = {10.1109/MSP.2005.1458289}
}

@article{alarifi2016uwb_indoor_review,
  title   = {Ultra Wideband Indoor Positioning Technologies: Analysis and Recent Advances},
  author  = {Alarifi, Abdulrahman and Al-Salman, Abdulmalik and Alsaleh, Mansour and Alnafessah, Abdullah and Al-Hadhrami, Sami and Al-Ammar, M. A. and Al-Khalifa, Hend S.},
  journal = {Sensors},
  volume  = {16},
  number  = {5},
  pages   = {707},
  year    = {2016},
  doi     = {10.3390/s16050707}
}

@article{maalek2016uwb_construction_accuracy,
  title   = {Accuracy assessment of ultra-wide band technology in locating dynamic resources in indoor scenarios},
  author  = {Maalek, Rami and Sadeghpour, Farid},
  journal = {Automation in Construction},
  volume  = {63},
  pages   = {12--26},
  year    = {2016},
  doi     = {10.1016/j.autcon.2015.11.009}
}

@article{ochoa2024uwb_roadworker_safety,
  title   = {Untethered Ultra-Wideband-Based Real-Time Locating System for Road-Worker Safety},
  author  = {Ochoa-de-Eribe-Landaberea, Aitor and Zamora-Cadenas, Leticia and Velez, Igone},
  journal = {Sensors},
  volume  = {24},
  number  = {8},
  pages   = {2391},
  year    = {2024},
  doi     = {10.3390/s24082391}
}

@article{wang2023uwb_nlos_survey,
  title   = {Survey on NLOS Identification and Error Mitigation for UWB Indoor Positioning},
  author  = {Wang, Fang and Tang, Hai and Chen, Jialei},
  journal = {Electronics},
  volume  = {12},
  number  = {7},
  pages   = {1678},
  year    = {2023},
  doi     = {10.3390/electronics12071678}
}

@article{angarano2021robust_uwb_deep_edge,
  title   = {Robust Ultra-wideband Range Error Mitigation with Deep Learning at the Edge},
  author  = {Angarano, Simone and Mazzia, Vittorio and Salvetti, Francesco and Fantin, Giovanni and Chiaberge, Marcello},
  journal = {Engineering Applications of Artificial Intelligence},
  volume  = {102},
  pages   = {104278},
  year    = {2021},
  doi     = {10.1016/j.engappai.2021.104278}
}

@article{wang2021semi_supervised_uwb_wcl,
  title   = {A Semi-Supervised Learning Approach for UWB Ranging Error Mitigation},
  author  = {Wang, Tianyu and Hu, Ke and Li, Zhenjie and Lin, Kai and Wang, Jinzhao and Shen, Yuan},
  journal = {IEEE Wireless Communications Letters},
  volume  = {10},
  number  = {3},
  pages   = {688--691},
  year    = {2021},
  doi     = {10.1109/LWC.2020.3046531}
}

@article{li2023semi_supervised_uwb_waveform,
  title         = {A Semi-Supervised Learning Approach for Ranging Error Mitigation Based on UWB Waveform},
  author        = {Li, Yuxiao and Mazuelas, Santiago and Shen, Yuan},
  journal       = {arXiv preprint arXiv:2305.18208},
  year          = {2023},
  eprint        = {2305.18208},
  archivePrefix = {arXiv}
}

@article{volpi2023lowcost_uwb_rtls,
  title   = {Low-Cost UWB Based Real-Time Locating System: Development, Lab Test, Industrial Implementation and Economic Assessment},
  author  = {Volpi, Andrea and Tebaldi, Letizia and Matrella, Guido and Montanari, Roberto and Bottani, Eleonora},
  journal = {Sensors},
  volume  = {23},
  number  = {3},
  pages   = {1124},
  year    = {2023},
  doi     = {10.3390/s23031124}
}

@article{qiyue2015adaptive_kf_nlos,
  title   = {A novel adaptive Kalman filter based NLOS error mitigation algorithm},
  author  = {Li, Qiyue and Wu, Zhong and Li, Jie and Sun, Wei and Wang, Jianping},
  journal = {IFAC-PapersOnLine},
  volume  = {48},
  number  = {28},
  pages   = {1118--1123},
  year    = {2015},
  doi     = {10.1016/j.ifacol.2015.12.281}
}

@inproceedings{zaheer2017deepsets,
  title     = {Deep Sets},
  author    = {Zaheer, Manzil and Kottur, Satwik and Ravanbakhsh, Siamak and Poczos, Barnabas and Salakhutdinov, Ruslan R and Smola, Alexander J},
  booktitle = {Advances in Neural Information Processing Systems},
  volume    = {30},
  pages     = {3391--3401},
  year      = {2017}
}

@inproceedings{lee2019set_transformer,
  title     = {Set Transformer: A Framework for Attention-based Permutation-Invariant Neural Networks},
  author    = {Lee, Juho and Lee, Yoonho and Kim, Jungtaek and Kosiorek, Adam and Choi, Seungjin and Teh, Yee Whye},
  booktitle = {Proceedings of the 36th International Conference on Machine Learning},
  series    = {Proceedings of Machine Learning Research},
  volume    = {97},
  pages     = {3744--3753},
  year      = {2019},
  publisher = {PMLR},
  url       = {https://proceedings.mlr.press/v97/lee19d.html}
}

@inproceedings{assran2023ijepa,
  title     = {Self-Supervised Learning from Images with a Joint-Embedding Predictive Architecture},
  author    = {Assran, Mahmoud and Duval, Quentin and Misra, Ishan and Bojanowski, Piotr and Vincent, Pascal and Rabbat, Michael and LeCun, Yann and Ballas, Nicolas},
  booktitle = {Proceedings of the IEEE/CVF Conference on Computer Vision and Pattern Recognition (CVPR)},
  pages     = {15619--15629},
  year      = {2023},
  url       = {https://openaccess.thecvf.com/content/CVPR2023/html/Assran_Self-Supervised_Learning_From_Images_With_a_Joint-Embedding_Predictive_Architecture_CVPR_2023_paper.html}
}

@article{li2025robotic,
  title={On the robotic uncertainty of fully autonomous traffic: From stochastic car-following to mobility--safety trade-offs},
  author={Li, Hangyu and Sun, Xiaotong and Zhuang, Chenglin and Li, Xiaopeng},
  journal={Transportation Research Part C: Emerging Technologies},
  volume={178},
  pages={105254},
  year={2025},
  publisher={Elsevier}
}

@inproceedings{ghosh2025roadwork,
  title={Roadwork: A dataset and benchmark for learning to recognize, observe, analyze and drive through work zones},
  author={Ghosh, Anurag and Zheng, Shen and Tamburo, Robert and Vuong, Khiem and Alvarez-Padilla, Juan and Zhu, Hailiang and Cardei, Michael and Dunn, Nicholas and Mertz, Christoph and Narasimhan, Srinivasa G},
  booktitle={Proceedings of the IEEE/CVF International Conference on Computer Vision},
  pages={6132--6142},
  year={2025}
}

@inproceedings{shi2021work,
  title={Work zone detection for autonomous vehicles},
  author={Shi, Weijing and Rajkumar, Ragunathan Raj},
  booktitle={2021 IEEE International Intelligent Transportation Systems Conference (ITSC)},
  pages={1585--1591},
  year={2021},
  organization={IEEE}
}

@article{niu2023deep,
  title={Deep learning-based ranging error mitigation method for UWB localization system in greenhouse},
  author={Niu, Ziang and Yang, Huizhen and Zhou, Lei and Taha, Mohamed Farag and He, Yong and Qiu, Zhengjun},
  journal={Computers and electronics in agriculture},
  volume={205},
  pages={107573},
  year={2023},
  publisher={Elsevier}
}

@article{yang2024ultra,
  title={Ultra-wideband ranging error mitigation with novel channel impulse response feature parameters and two-step non-line-of-sight identification},
  author={Yang, Hongchao and Wang, Yunjia and Xu, Shenglei and Bi, Jingxue and Jia, Haonan and Seow, Cheekiat},
  journal={Sensors},
  volume={24},
  number={5},
  pages={1703},
  year={2024},
  publisher={MDPI}
}

@article{liu2024adaptive,
  title={Adaptive optimization strategy and evaluation of vehicle-road collaborative perception algorithm in real-time settings},
  author={Liu, Jiaxi and Gao, Bolin and Zhong, Wei and Lu, Yanbo and Han, Shuo},
  journal={Computers and Electrical Engineering},
  volume={120},
  pages={109785},
  year={2024},
  publisher={Elsevier}
}

\end{document}